\pdfoutput=1
\documentclass[11pt]{article}

\usepackage[]{acl}

\usepackage{enumitem}
\usepackage{microtype}
\usepackage{makecell}
\usepackage{inconsolata}
\usepackage{colortbl}

\usepackage{times}
\usepackage{latexsym}
\usepackage{amsmath}

\usepackage{times}
\usepackage{latexsym}
\usepackage[T1]{fontenc}

\usepackage[T1]{fontenc}
\usepackage[utf8]{inputenc}

\usepackage{microtype}

\usepackage{inconsolata}

\usepackage{graphicx}
\usepackage{amsfonts}
\usepackage{amsmath}

\usepackage{booktabs}
\usepackage{multirow}
\usepackage{amsmath}
\usepackage{microtype}
\usepackage{graphicx}
\usepackage{amsfonts}
\usepackage{tabularx}
\usepackage{color}
\usepackage{comment}
\usepackage{amsmath,amsfonts,amssymb}
\usepackage{arydshln}
\usepackage{tablefootnote}
\usepackage{xcolor}
\usepackage{subcaption}
\usepackage{float}

\usepackage{bbding}
\usepackage{pifont}
\usepackage{wasysym}
\usepackage{utfsym}
\usepackage{fontawesome}
\usepackage[most]{tcolorbox}

\usepackage{amsthm}

\title{Document Reconstruction Unlocks Scalable Long-Context RLVR}

\author{
Yao Xiao$^{1,2}$
Lei Wang$^{1}$
Yue Deng$^{1}$
Guanzheng Chen$^{1}$
Ziqi Jin$^{1}$\\
\textbf{
Jung-jae Kim$^{3}$
Xiaoli Li$^{2}$
Roy Ka-wei Lee$^{2}$
Lidong Bing$^{1}$}
\\
$^1$Infinity Lab, MiroMind AI
$^2$Singapore University of Technology and Design\\
$^3$Institute for Infocomm Research, A*Star, Singapore\\
\\
}

\begin{document}
\maketitle

\begin{abstract}

Reinforcement Learning with Verifiable Rewards~(RLVR) has become a prominent paradigm to enhance the capabilities (i.e.\ long-context) of Large Language Models~(LLMs).
However, it often relies on gold-standard answers or explicit evaluation rubrics provided by powerful teacher models or human experts, which are costly and time-consuming.
In this work, we investigate unsupervised approaches to enhance the long-context capabilities of LLMs, \emph{eliminating the need for heavy human annotations or teacher models' supervision}.
Specifically, we first replace a few paragraphs with special placeholders in a long document.
LLMs are then trained through reinforcement learning to reconstruct the document by correctly identifying and sequencing missing paragraphs from a set of candidate options.
This training paradigm enables the model to capture global narrative coherence, significantly boosting long-context performance. 
We validate the effectiveness of our method on two widely used benchmarks, RULER and LongBench~v2.
While acquiring noticeable gains on RULER (nearly 10 points), it can also achieve a reasonable improvement on LongBench~v2 without any manually curated long-context QA data.
Furthermore, we conduct extensive ablation studies to analyze the impact of reward designs, data curation strategies, training schemes, and data scaling effects on model performance.
We release our code, data, and models\footnote{\url{https://github.com/XYaoooo/reconstruction_long}}.

\end{abstract}

\section{Introduction}

Reinforcement Learning with Verifiable Rewards (RLVR) has recently achieved the state-of-the-art in Large Language Model (LLM) reasoning~\citep{kumar2025training, yu2025dapoopensourcellmreinforcement, yeo2025demystifying, zeng2025simplerlzoo, liu2025prorl, claude_ai}.
Using ground-truth feedback to guide generation, RLVR enables models like DeepSeek-R1~\citep{Guo_2025} to navigate complex multi-step problem-solving paths in domains such as mathematics and programming with unprecedented precision~\citep{yang2025qwen3technicalreport}. 
However, as LLMs evolve into agents that must interact with expansive real-world datasets, the challenge shifts from local reasoning to global context processing~\citep{miromind2025mirothinker, tongyideepresearchteam2025tongyideepresearchtechnicalreport, prabhakar2025enterprisedeepresearch}. 
Here, a significant gap remains: models that excel at step-by-step logic reasoning often struggle to maintain coherence when retrieving and synthesizing information across tens of thousands of tokens~\citep{ wu2025retrieval, zhuang2025scaling, wu2025longwriterzeromasteringultralongtext, bai2025longwriter}. 
This suggests that reasoning gains do not necessarily scale with context length, leaving a gap between reasoning capacity and long-context understanding.

\begin{figure}[t]
\centering
\includegraphics[width=0.95\linewidth]{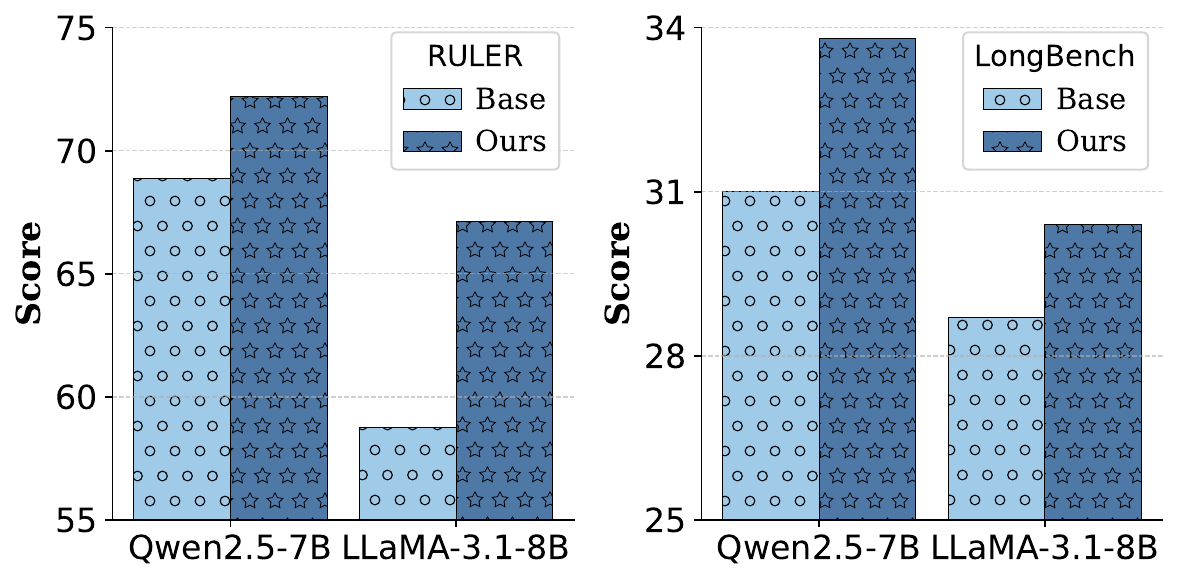}
\caption{
Average score on RULER and overall score on LongBench~v2
for Qwen2.5-7B-Instruct-1M and LLaMA-3.1-8B-Instruct.
}
\label{fig:performance_comp}
\vspace{-1.5em}
\end{figure}

The pursuit of extending the context window has thus become a central objective in developing frontier LLMs~\citep{zhu2024pose, an2024trainingfree, peng2024yarn, gao-etal-2025-train, lu2025mobamixtureblockattention, yang2025qwen251mtechnicalreport, wan2025qwenlongl1longcontextlargereasoning, xu2026alternatingreinforcementlearningrubricbased}.
From analyzing massive code bases to distilling insights from an entire document, the ability to process and reason over long sequences is essential for practical applications. 
However, as the context length increases, the difficulty of maintaining global coherence and performing precise retrieval increases exponentially. 
Models frequently suffer from the well-documented ``lost in the middle''  phenomenon~\citep{liu-etal-2024-lost}, where information in the center of a long prompt is ignored, or they struggle to maintain the logical consistency of a narrative over tokens~\citep{hsieh-etal-2024-found, hsieh2024ruler, du-etal-2025-context}.
Although RLVR offers a powerful framework for LLMs to refine their long-range dependency handling, its application is currently restricted by a heavy reliance on external supervision.

Recent RLVR approaches~\citep{wang2025loongrlreinforcementlearningadvanced, chen2026longrlvr} to enhance the
long-context capability of LLMs are constrained by the availability of gold-standard answers or evaluation rubrics, which are typically provided by expensive human experts and frontier teacher models (often close-source)~\citep{chen2025longpo, zhang-etal-2025-longreward, huang2025reinforcementlearningrubricanchors}. 
This dependency creates a significant scalability bottleneck: while the need for long-context understanding is universal, question-answering pairs from human annotations required for training are prohibitively expensive and difficult to generate at the scale needed for reinforcement learning. 
Furthermore, relying on teacher models can introduce biases or limit the potential of student models to the capabilities of the supervisor~\citep{kim2025rlvr, cheng2025do}. 
To unlock the next level of long-context ability of LLMs, we explore a training mechanism that can derive objective, verifiable rewards directly from the data itself in an unsupervised manner.

In this work, we introduce a fully unsupervised RLVR framework that bypasses the need for costly human annotations or rubrics from teacher models, enabling a more scalable approach to long-context training.
We hypothesize that long documents possess an inherent structure, specifically their narrative flow and logical sequence, which contains valuable internal signals. 
The signals can act as a natural and verifiable reward for training model through RLVR. 
We formalize this through a document reconstruction task.
Specifically, we first mask a few random paragraphs within a long document and then require the LLM to correctly identify and sequence these missing segments from a shuffled pool of candidates.
Since the original documents provide the ground truth, the resulting reward enjoys the desirable property of being both verifiable and fully automated.
This reconstruction training mechanism encourages LLMs to move beyond surface-level pattern matching and instead develop a deeper, more structural understanding of the global context~\citep{NEURIPS2019_dc6a7e65}.

We evaluate our method on two of the most rigorous benchmarks in the field: RULER~\citep{hsieh2024ruler} and LongBench~v2~\citep{bai-etal-2025-longbench}.
Our empirical results demonstrate that this unsupervised paradigm yields substantial gains on RULER and moderate improvement performance on LongBench~v2, as shown in Figure~\ref{fig:performance_comp}. 
These findings suggest that the underlying long-range structural integrity of documents provides a valuable training paradigm, offering a scalable path toward more capable long-context LLMs.
Our contributions are summarized as follows:

\begin{itemize}

\item We propose an unsupervised RLVR framework based on document reconstruction and formulate it as a sequential selection problem with verifiable rewards, encouraging models to learn global narrative coherence and long-range structural dependencies.
\item We validate our approach through extensive experiments on RULER and LongBench~v2, showing that it provides a scalable and effective alternative to supervised long-context training.
\item We perform comprehensive ablation studies to examine the effects of reward formulation, data curation strategies, and training configurations, offering deeper insights into the characteristics of our method.
\end{itemize}

\section{Background}

\paragraph{Group Relative Policy Optimization (GRPO).}
Group Relative Policy Optimization (GRPO)~\citep{shao2024deepseekmathpushinglimitsmathematical, Guo_2025} 
is a policy-gradient method that removes the need for an explicit value function by estimating advantages through relative comparisons within a group of sampled responses.
Given a query \( q \), GRPO samples a group of trajectories \( \{ \tau^{(i)} \}_{i=1}^{G} \sim \pi_{\theta_{\text{old}}}(\cdot \mid q) \) and assigns each trajectory a scalar reward \( R_{\tau^{(i)}} \).
The advantage for each trajectory is computed by normalizing its reward against the group:
\[
A(\tau^{(i)}) =
\frac{
R_{\tau^{(i)}} - \operatorname{mean}\!\left( \{ R_{\tau^{(j)}} \}_{j=1}^{G} \right)
}{
\operatorname{std}\!\left( \{ R_{\tau^{(j)}} \}_{j=1}^{G} \right)
}
\]
GRPO then optimizes a PPO-style clipped surrogate objective using these group-relative advantages, enabling stable policy updates without learning a separate value function.

Using the normalized group-relative advantages, GRPO performs policy updates via a PPO-style clipped surrogate objective:
\begin{equation}
\mathcal{L}_{\text{GRPO}}(\theta)
=
\mathbb{E}_{\tau}
\Big[
\min \big(
r_\theta A,
\text{clip}(r_\theta, 1-\epsilon, 1+\epsilon) A
\big)
\Big]
\end{equation}
where
\[
r_\theta(\tau)
= \frac{\pi_\theta(\tau \mid q)}{\pi_{\theta_{\text{old}}}(\tau \mid q)}.
\]
This clipped objective constrains policy updates while leveraging group-normalized rewards as a low-variance advantage estimator, eliminating the need for a learned critic.

\section{Method}

\label{main_method}

\begin{figure*}[t]
    \centering
    \includegraphics[
        width=0.85\linewidth,
        trim=2.5cm 6.5cm 2.5cm 3.5cm]{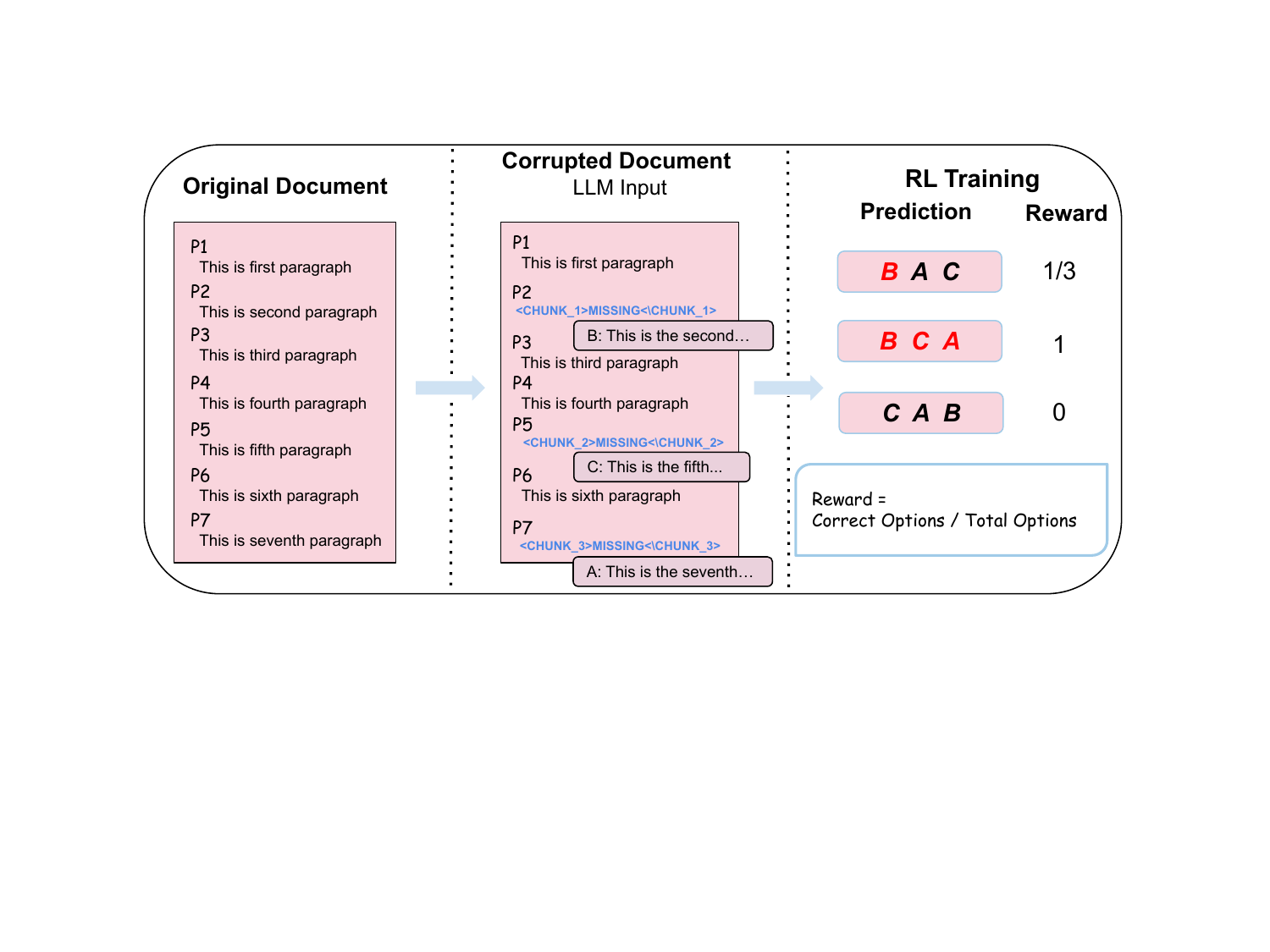}
    \caption{Overview of the document reconstruction framework.
    Given a long document, we corrupt it by selecting some paragraphs and shuffle them as options. We train LLMs via RLVR to reconstruct the document by generating the option sequence by order.}
    \label{fig:overview}
    \vspace{-1.51em}
\end{figure*}

\subsection{Task Formulation: Document Reconstruction}

Our goal is to derive a verifiable reward from raw documents without labels from human annotators. 
Given a long document consisting of $n$ paragraphs, denoted as $D = \{p_1, p_2, \ldots, p_n\}$, we first select a subset of paragraphs to mask.
The original document is then transformed into a corrupted context, where the masked paragraphs are replaced by placeholders marked with identifiers, denoted as \texttt{<CHUNK\_i>MISSING</CHUNK\_i>}.
$i$ means the $i$-th masked paragraph of the document.

The model is presented with the context and a set of shuffled candidates labeled with options. 
LLMs are asked to reconstruct the original text by first reasoning and then generating a list of these options in the correct order. 
For example, if four paragraphs were selected, the model's output would be a formatted list such as $\{B, A, D, C\}$. 
This formulation transforms the long context understanding problem into a sequential decision making task, where the model must utilize global narrative flow and logical consistency to determine the correct placement of each segment.
The overview of our method can be found in Figure~\ref{fig:overview}.

\subsection{Reward Design}

Unlike open-ended generation, our reconstruction task provides an objective and verifiable answer. 
Since the original document provides ground truth ordering, we define a verifiable reward function that evaluates the model output against ground truth as follows:

\begin{equation}
R(o, g) =
\begin{cases}
1, & \text{if } o = g \\[8pt]
\frac{1}{K} \sum_{i=1}^K \mathbb{I}[o_i = g_i], & \begin{aligned} \text{if } & \mathcal{V}(o) \\ & \land o \neq g \end{aligned} \\[16pt]
0, & \text{otherwise}
\end{cases}
\end{equation}

In this formulation, $K$ represents the total number of masked segments, while $o_i$ and $g_i$ denote
the predicted and ground-truth options for the $i$-th placeholder, respectively.
The function $\mathbb{I}[\cdot]$ is an indicator function that yields $1$ for a correct match and $0$ otherwise.
A critical component of this reward structure is the $\mathcal{V}(o)$ condition.
We define a predicted output as a \emph{valid permutation} if and only if the set of options provided in
the model's response is identical to the set of ground-truth options.
This constraint requires the model to correctly identify and utilize the complete pool of candidates, without omitting or duplicating any options~\citep{wu2025visualjigsawposttrainingimproves, lu2026goldengoosesimpletrick}.

This piecewise reward structure balances global accuracy with fine-grained feedback. 
A full reward of 1 is assigned when the reconstruction is exact ($o=g$). 
For outputs that are not perfectly ordered but remain structurally valid, defined as sequences whose predicted option set exactly matches the ground truth, we assign a partial reward proportional to the fraction of correctly placed segments. 
This design encourages progressive refinement of the global structure, even without full sequence accuracy.
In contrast, any output that violates the required format or constitutes an invalid permutation receives a reward of 0. 
The entire training signal is fully automated and does not depend on human annotations or external teacher models.

\subsection{Curriculum through Complexity Scaling}

An appealing advantage of our framework is the ability to precisely calibrate the difficulty of training samples by adjusting the complexity of the reconstruction task. 
Our intuition is that larger $K$ leads to more challenging samples.
As $K$ increases, the search space expands exponentially because the number of possible permutations for the candidate set is $K!$.
By treating $K$ as a tunable hyperparameter, we can implement a curriculum~\citep{Bengio2009CurriculumL} that allows the model to first master local coherence with fewer options before tackling the complex global structural dependencies required for extremely long contexts. 
This controllable difficulty~\citep{zeng2025rlvescalingreinforcementlearning, wang2026rlanythingforgeenvironmentpolicy} ensures that the model can progressively build its long-context understanding capabilities.
\section{Experimental Setup}

\paragraph{Data Curation.}
We source our training documents from the corpus provided by \citet{chen2025longpo}, which covers three diverse domains: books, arXiv papers, and code. 
We curate a subset consisting of the 8,000 longest documents from the book domain, along with 3,000 longest documents each from arXiv and code~(total 14,000).
We apply varying levels of difficulty to these documents by setting $K \in \{2, 4, 6, 8\}$ with their corresponding number ratio being $3:3:3:5$, which we find useful in our preliminary experiment.
The average token number of samples in the training set is 49,000.
This multi-scale approach to the number of masked segments ensures that the training set provides a broad spectrum of structural challenges.
In addition, we also curate 500 samples for validation set to better observe training dynamics.

\paragraph{Training.}
We adopt the curriculum schedule~\citep{Bengio2009CurriculumL} to progressively train models by increasing $K$ from 2 to 8.
Our implementation is built on the Verl framework \citep{sheng2024hybridflow}. 
To optimize the training process, we employ Group Relative Policy Optimization (GRPO)~\citep{shao2024deepseekmathpushinglimitsmathematical}.
For RL training, we use the AdamW optimizer with a constant learning rate of 1e-6 and a 5-step linear warmup. 
For rollout, we use a prompt batch size of 128 and sample 8 responses per prompt, with a maximum context length of 64K and a response length of 4096.
Our reconstruction prompt can be found in Appendix~\ref{reprompt}.

\paragraph{Evaluation.}
Benchmarks. We evaluate all models on two challenging long-context QA benchmarks: 
(1) RULER~\citep{hsieh2024ruler}: A synthetic benchmark testing multi-hop reasoning over arbitrary context length.
Specifically, we evaluate on four tasks of it (Variable Tracking, Frequent Words
Extraction, Common Words
Extraction, Question Answering).
(2) LongBench~v2~\citep{bai-etal-2025-longbench}: A realistic multi-choice QA benchmark on documents up to 128K tokens.
We evaluate the lengths of 32K, 64K, and 128K.
In our analysis, we primarily focus on RULER, as it provides results of a more comprehensive and diverse task.

\paragraph{Models and Baselines.}
We select LLaMA-3.1-8B-Instruct and Qwen2.5-7B-Instruct-1M as our backbone models, which also serve as our baseline.
A discussion about continual pretraining on long documents is also presented in Appendix~\ref{continue}.

\section{Results and Analysis}

\subsection{Main Results}

\begin{figure*}[t]
    \centering

    \begin{subfigure}{\linewidth}
        \centering
        \includegraphics[width=0.88\linewidth]{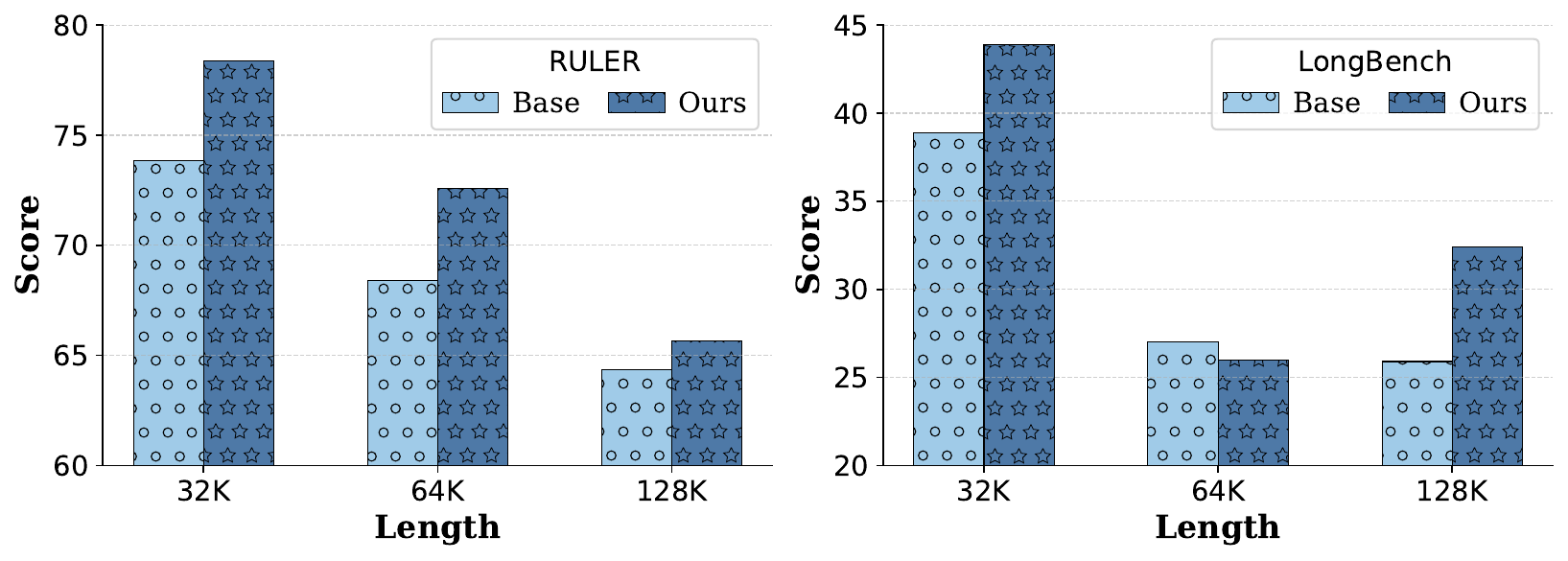}
        \caption{Qwen2.5-7B-Instruct-1M}
        \label{fig:ruler}
    \end{subfigure}

    \begin{subfigure}{\linewidth}
        \centering
        \includegraphics[width=0.88\linewidth]{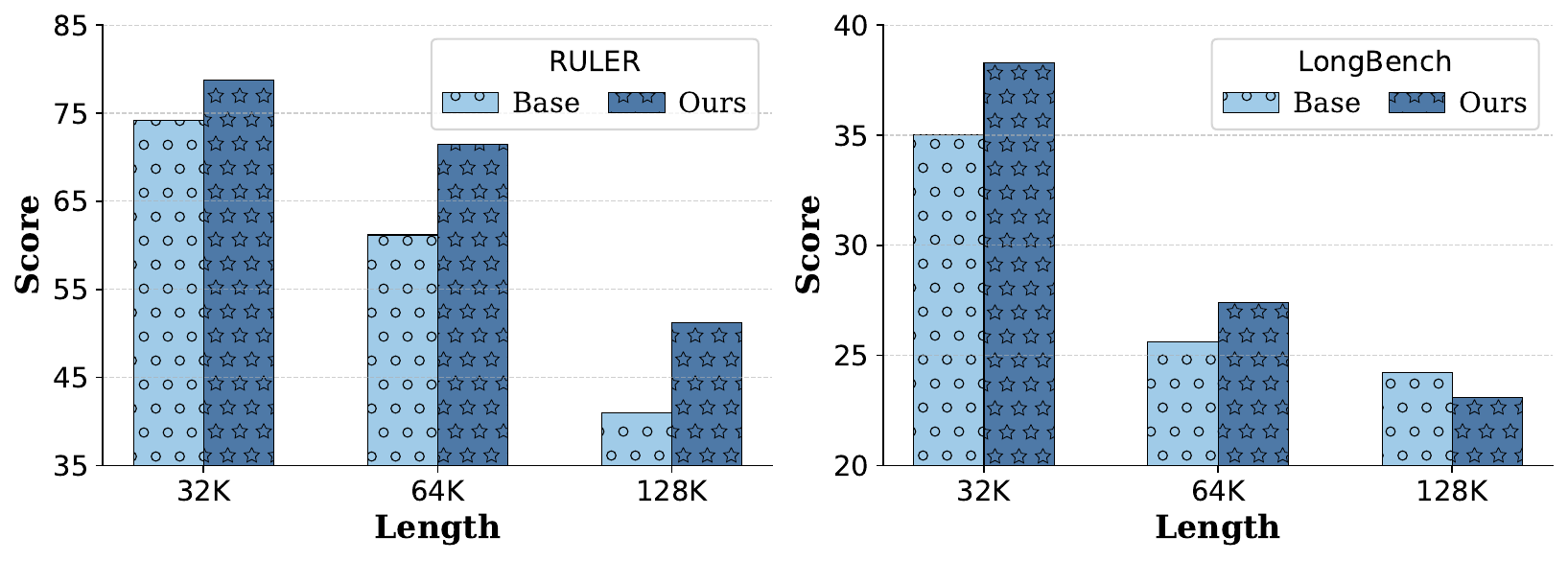}
        \caption{LLaMA-3.1-8B-Instruct}
        \label{fig:longbench}
    \end{subfigure}

    \caption{Performance comparison across different context lengths and models.}
    \label{fig:main_res}
\end{figure*}
 
We summarize the main results across different context lengths and backbone models in Figure~\ref{fig:main_res}. 
In addition, we also report the average score of RULER and the overall score of LongBench~v2 in Figure~\ref{fig:performance_comp}.

On RULER, our method produces substantial improvements, with consistent gains as the context length increases from 32K to 128K.
This indicates that our unsupervised reconstruction training effectively enhances the model’s ability to maintain global coherence and retrieve relevant information in long contexts.
On LongBench~v2, we also observe moderate improvements across most context lengths. 
And these gains are achieved without using any manually curated long-context QA data, highlighting the effectiveness of reconstruction training even for question answering. 
The improvements remain consistent across different backbone architectures, including Qwen2.5-7B-Instruct-1M and LLaMA-3.1-8B-Instruct, demonstrating that our method is not tied to a specific model family.

\begin{table}[t]
  \centering
  \small
  \vspace{0.4em}
  \begin{tabular}{lccc}
    \toprule
    \textbf{Model} & \textbf{\# Data} & \textbf{RULER-QA} & \textbf{LongBench} \\
    \midrule
    \multicolumn{4}{l}{\textbf{Qwen2.5-7B-Instruct-1M}} \\
    \quad SFT  & 46K & 64.5 & 33.2 \\
    \quad Ours & 14K & \textbf{68.3} & \textbf{33.8} \\
    \midrule
    \multicolumn{4}{l}{\textbf{LLaMA-3.1-8B-Instruct}} \\
    \quad SFT  & 46K & \textbf{64.7} & \textbf{31.2} \\
    \quad Ours & 14K & 61.2 & 30.4 \\
    \bottomrule
  \end{tabular}
  \caption{Comparison between SFT and our reconstruction training on RULER-QA (average) and LongBench v2 (overall).}
  \label{tab:sftcomp}
  \vspace{-1.5em}
\end{table}

Taken together, these results show that unsupervised document reconstruction via RLVR provides a scalable and effective mechanism for improving long-context capabilities. 
The method delivers strong gains on synthetic reasoning benchmarks and meaningful improvements on realistic QA tasks, all while eliminating the need for human annotations or teacher-model supervision.
Detailed scores on RULER can be found in Appendix~\ref{reuler_app}.
We also record the performance on our curated validation set during the training process in Appendix~\ref{valid}.

\paragraph{Comparison to SFT}
\cite{chen2026longrlvr} employ a powerful teacher model to curate 46K context-specific QA pairs for supervised fine-tuning. 
They report results on LongBench v2 and only the QA subset of RULER.
In Table~\ref{tab:sftcomp}, we compare our results with theirs. 
With only 14K training samples and without relying on context-specific QA pairs, our approach achieves better performance on Qwen2.5-7B-Instruct-1M, while slightly lags behind on LLaMA-3.1-8B-Instruct.

\subsection{Dense vs.\ Sparse Reward}

In our main experiments, we employ a dense reward that provides partial credit for partially correct reconstructions. 
To better understand the role of reward shaping in document reconstruction, we compare this design with a sparse reward formulation, defined as

\begin{equation}
R(o, g) =
\begin{cases}
1, & \text{if } o = g, \\[4pt]
0, & \text{otherwise}.
\end{cases}
\end{equation}

\begin{figure}[t]
  \centering
  \includegraphics[width=0.38\textwidth]{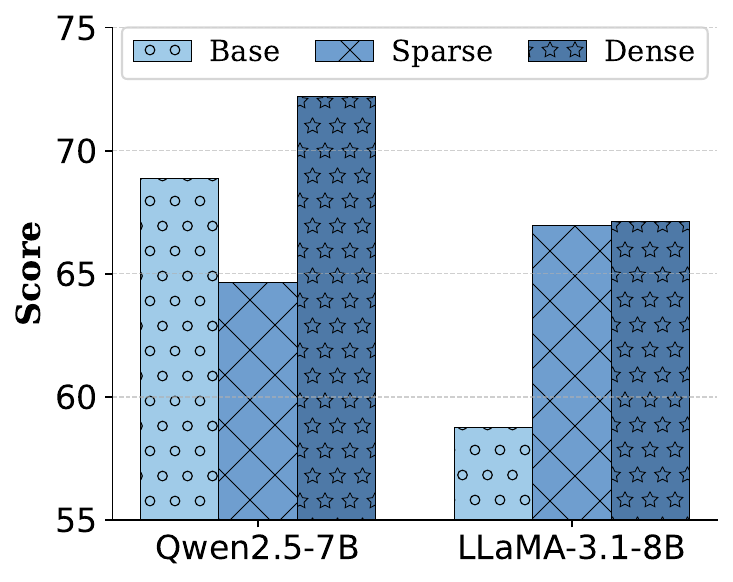}
  \caption{Average scores on RULER. We compare the performance of dense and sparse rewards.}
  \label{fig:dense_sparse}
\end{figure}

This sparse reward assigns a non-zero signal only when the predicted ordering exactly matches the ground truth, providing a stricter but less informative supervision signal.
As shown in Figure~\ref{fig:dense_sparse}, sparse rewards obtain performance similar to dense rewards on LLaMA-3.1-8B-Instruct.
However, it causes significant performance degradation on Qwen2.5-7B-Instruct-1M.
We hypothesize that sparse rewards are more likely to cause training instability due to the sparsity of positive rewards in the training process.

\subsection{Robustness to Option Mixture Ratios}

\begin{figure}[t]
  \centering
  \includegraphics[width=0.38\textwidth]{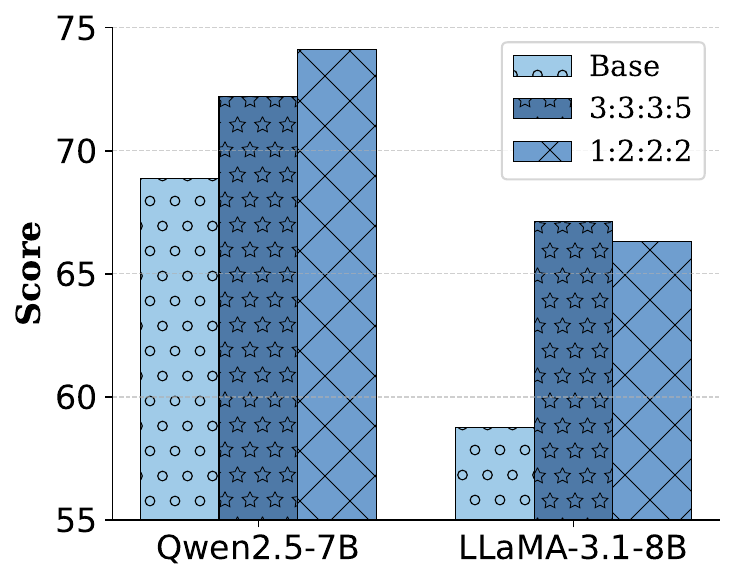}
  \caption{Average scores of RULER.
  We compare the performance of different option length mixture ratios.}
  \label{fig:ratio}
\end{figure}

In our main experiments, we adopt the option mixture ratio for $K = 2, 4, 6,$ and $8$ as $3:3:3:5$, which assigns only a moderate portion of training samples to small values of $K$ (e.g., $K$=2). 
To further validate robustness, we conduct an ablation study by shifting more training samples toward larger $K$, using the ratio $1:2:2:2$. 
In Figure~\ref{fig:ratio}, empirical results demonstrate that our method maintains high performance across varying option length distributions. 
For the Qwen2.5-7B-Instruct-1M model, the $1:2:2:2$ ratio yielded the highest score, surpassing baseline. 
In the case of LLaMA-3.1-8B-Instruct model, the $3:3:3:5$ ratio proves most effective.
These findings suggest that while specific ratios can offer marginal gains on different model architecture, the overall framework remains robust. 
The consistency in performance across different difficulty blends confirms that the method does not rely on a brittle or overly specific data composition to succeed.

\subsection{Longer Documents Bring Consistent Improvement}

During data curation of main experiments, we retain the longest 8,000, 3,000, and 3,000 documents from the book, arXiv, and code domains, respectively. 
In this experiment, we compare this document selection strategy against two alternative counterparts: shortest and random. 
The shortest counterpart selects the shortest documents while preserving the same domain ratios and total number of documents. 
The random counterpart samples documents randomly, keeping all other factors identical.

\begin{figure}[t]
 
  \centering
  \includegraphics[width=0.42\textwidth]{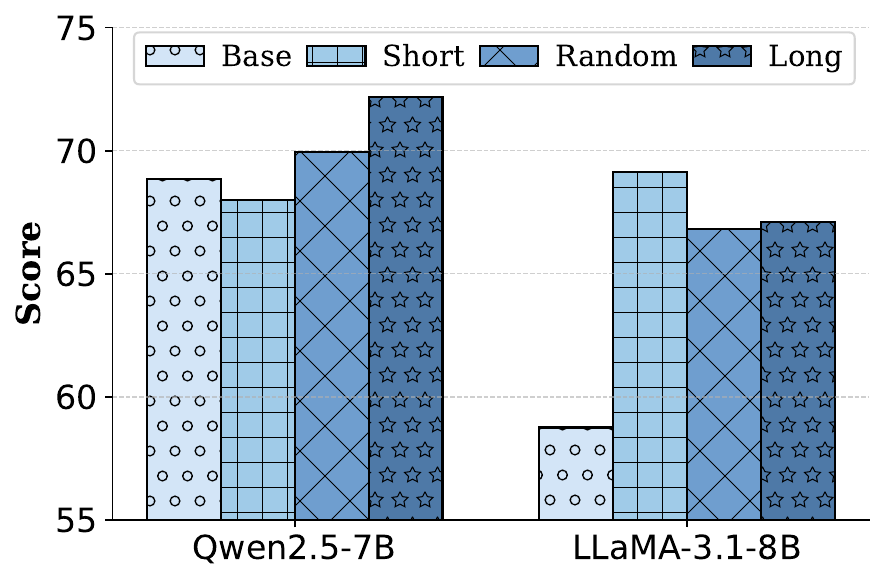}
  \caption{We report the average score of RULER. }
  \label{fig:length_comp}
  \vspace{-1em}
\end{figure}

As shown in Figure~\ref{fig:length_comp}, although short-document training brings modest gains for LLaMA-3.1-8B-Instruct, it does not exceed the baseline performance of Qwen2.5-7B-Instruct-1M.
Moreover, random document sampling leads to consistent but limited improvements on Qwen2.5-7B-Instruct-1M. 
In contrast, training on long documents leads to overall improvements, suggesting the importance of longer contexts for the reconstruction task. 
In summary, these observations indicate that document length plays a critical role in enabling effective long-context understanding through reconstruction training.

\subsection{More Documents, Better Performance}

\begin{figure}[t]
  \centering
  
  \includegraphics[width=0.33\textwidth]{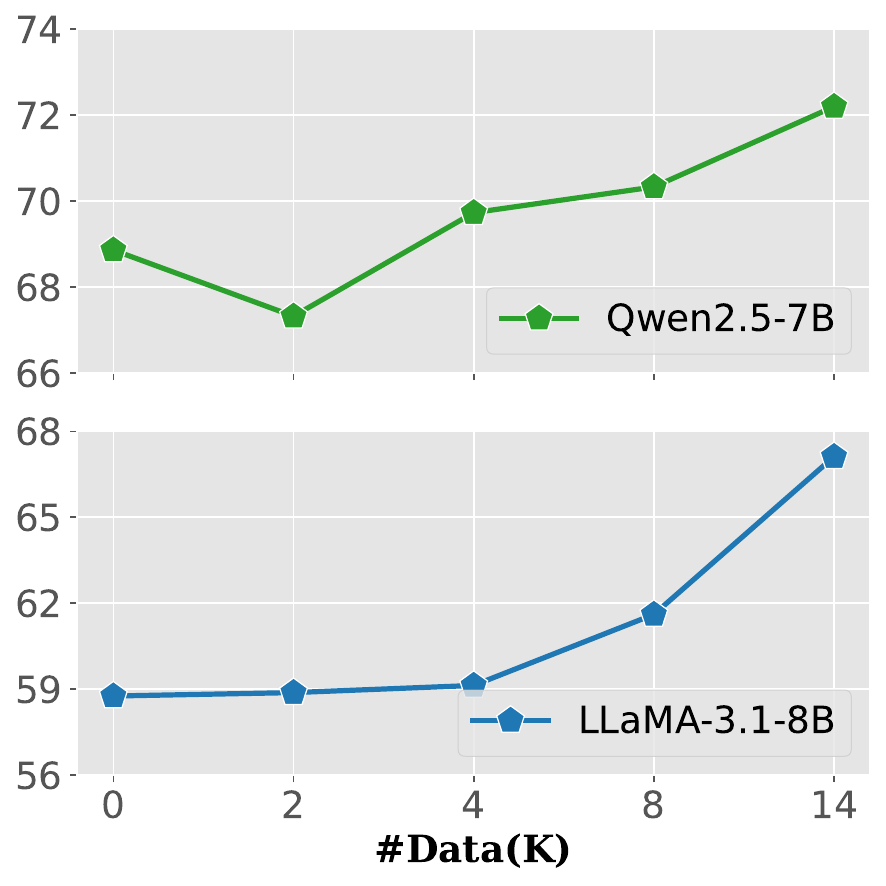}
  \caption{Average scores of RULER when we scale data from 0 to 14K.}
  \label{fig:scale_data}
  \vspace{-2em}
\end{figure}

Scaling training data is an important factor in understanding the effectiveness of reconstruction learning. 
In this part, we study the effect of training data scale on model performance by increasing the number of reconstruction training samples from 0 to 14,000. 
This setting allows us to examine how model performance evolves when we increase the number of reconstruction samples.
For each data scale, we train models under identical optimization settings and evaluate them on RULER. 
When the data size is set to 0, the model corresponds to the original backbone without reconstruction training, serving as a baseline.
We record the performance in Figure~\ref{fig:scale_data}, model performance generally improves as the number of reconstruction training samples increases.

Although performance may fluctuate slightly at smaller data scales, especially when the training set is limited, the overall trend is clearly positive as more data is introduced.
In particular, performance consistently increases when the data scale exceeds 4,000 samples, indicating that sufficient reconstruction data is crucial for effectively enhancing long-context understanding. 
These results suggest that reconstruction-based training scales well with data size and that increasing training data is an effective and stable way to improve model performance.
Notably, we do not observe a clear performance plateau within the evaluated data range, suggesting that the model may continue to benefit from additional reconstruction training data.
\textbf{Significant performance gains} (nearly 10 points) can be achieved on Qwen2.5-7B-Instruct-1M by scaling data to 3,0000, which is shown in Table~\ref{fig:scale_3w}.
Data recipe about this can be found in Appendix~\ref{scale3w}.

\begin{figure}[t]

  \centering
  \includegraphics[width=0.375\textwidth]{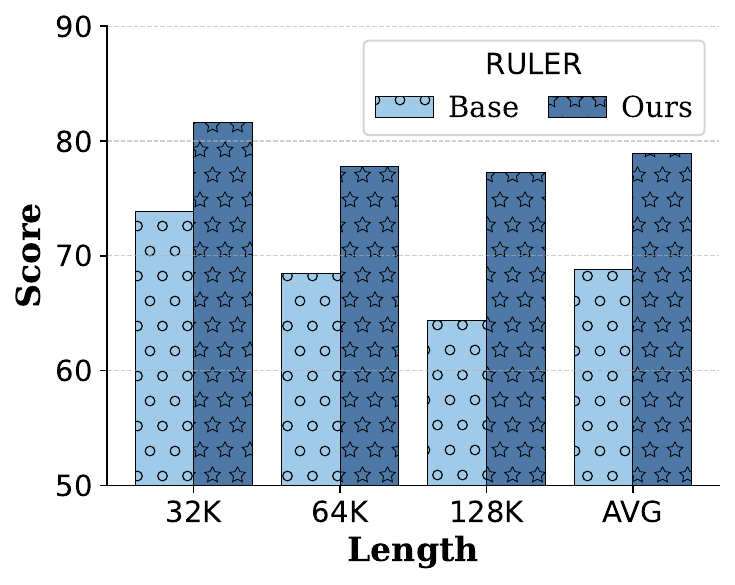}
  \caption{We report the performance gains (nearly 10 points) of RULER when scaling the data to 30,000 samples on Qwen2.5-7B-Instruct-1M.}
  \label{fig:scale_3w}
  \vspace{-1em}
\end{figure}

\subsection{Training Strategy: Shuffle or Not}

As mentioned in Section~\ref{main_method}, we curate training data by progressively increasing the option length $K$ from 2 to 8, following a curriculum-style training strategy. 
Our intuition is that reconstruction tasks with larger $K$ are substantially more challenging, and gradually exposing the model to harder samples can facilitate more stable optimization.

To validate this design choice, we compare the curriculum strategy with a shuffled training variant.
As shown in Table~\ref{tab:training}, the curriculum-based strategy (shuffle is false) consistently outperforms shuffled training for both Qwen2.5-7B-Instruct-1M and LLaMA-3.1-8B-Instruct backbones.
These results demonstrate that a curriculum of option length is an important component of our training framework.

\subsection{Analysis of Option Length}

In this part, we study the influence of sample difficulty on model performance.
Intuitively, a sample with more candidate options is more challenging, as the model must search over a larger permutation space and rely more heavily on global contextual cues.
We randomly sample 5,000 documents from the training corpus and create four distinct training sets, where each set is constructed using a fixed option length $K \in \{2, 4, 6, 8\}$, without mixing different values of $K$ within the same set.
We train an independent model for each curated training set, where each model is associated with a specific $K$.

\begin{table}[t]
  \centering
  \small
  \begin{tabular}{lcc}
    \toprule
    \textbf{Model} & \textbf{Qwen2.5-7B} & \textbf{LLaMA-3.1-8B} \\
    \midrule
    Base & 68.86 & 58.76 \\
    \midrule
    Shuffle \\
    \quad True  & 69.82 & 65.00 \\
    \quad False & \textbf{72.20} & \textbf{67.12} \\
    \bottomrule
  \end{tabular}
  \caption{Effect of training strategy during reconstruction training on RULER.}
  \label{tab:training}
  \vspace{-1em}
\end{table}

As shown in Table~\ref{tab:k_comparison}, we observe that model performance is relatively stable across different option lengths $K$. 
While increasing $K$ substantially enlarges the permutation space and task difficulty, it does not lead to a significant degradation in downstream RULER performance for either backbone. 
This suggests that the reconstruction-based training objective encourages robust global structure understanding rather than overfitting to a specific difficulty level.
Interestingly, intermediate option lengths ($K = 4$ and $K = 6$) slightly outperform both smaller and larger values of $K$, indicating a potential trade-off between task difficulty and learning efficiency.
In addition, training of single $K$ benefits weaker LLaMA-3.1-8B-Instruct, while degrades performance of stronger Qwen2.5-7B-Instruct-1M, suggesting different sensitivity to task specialization across backbones.
Qwen2.5-7B-Instruct-1M may be more likely to overfit to a single pattern of reconstruction.
It also highlights the importance our data curation strategy in main experiment, which mixes training samples from various $K$.

\section{Related Work}
\paragraph{Reinforcement Learning with Verifiable Rewards.}

Reinforcement Learning with Verifiable Rewards provides an objective and scalable framework for improving the reasoning capabilities of large language models by supervising them with ground-truth answers~\citep{lambert2025tulu3pushingfrontiers, Guo_2025}.
% This paradigm has proven particularly effective in domains with unambiguous correctness criteria, including mathematics and programming. 
Previous work shows that RLVR can elevate models to expert-level reasoning performance by encouraging the discovery of correct internal reasoning trajectories through outcome-only rewards~\citep{openai2024openaio1card, kim2025rlvr, huang2025winninggoldimo2025, mo2025multiagenttoolintegratedpolicyoptimization, wang2026visgymdiversecustomizablescalable}.
Most existing RLVR studies focus on self-contained reasoning tasks, where the central challenge is to recover a valid trajectory~\citep{yue2025doesreinforcementlearningreally}. 
In these settings, emergent behaviors, such as self-reflection, have been identified as important contributors to performance gains~\citep{gandhi2025cognitivebehaviorsenableselfimproving}. 
Recent reasoning-oriented models, including OpenAI’s o1 series and DeepSeek, have popularized RLVR-based training pipelines and reinforcement learning algorithms such as GRPO.

\begin{table}[t]
  \centering
  
  \begin{tabular}{lcc}
    \toprule
    \textbf{Model} & \textbf{Qwen2.5-7B} & \textbf{LLaMA-3.1-8B} \\
    \midrule
    Base  & 68.86 & 58.76 \\
    \midrule
    $K=2$ & 64.79 & 63.96 \\
    $K=4$ & 65.83 & 64.27 \\
    $K=6$ & 64.69 & 64.19 \\
    $K=8$ & 64.83 & 63.53 \\
    \bottomrule
  \end{tabular}
  \caption{Average performance of different $K$ values on RULER.}
  \label{tab:k_comparison}
  \vspace{-1em}
\end{table}

\paragraph{Long-Context Training of LLMs.}

Training large language models for long-context reasoning poses challenges that differ fundamentally from those addressed by standard reinforcement learning with verifiable rewards. 
The outcome reward provides limited guidance in long-context settings, where success depends on identifying and grounding relevant evidence from extensive context~\citep{wan2025qwenlongl1longcontextlargereasoning}.
Most existing approaches to long-context training rely on supervised fine-tuning with synthetic data rather than reinforcement learning~\citep{li-etal-2024-making, yen2025helmet, chen2025on}.
Prior work pads questions with unrelated passages, shuffles document order, or fills contexts with irrelevant text to artificially increase sequence length~\citep{trivedi-etal-2023-interleaving}. 
% However, synthetic-data construction often introduces superficial complexity and fails to produce high-quality and challenging examples, limiting the effectiveness of the resulting long-context capabilities.
Improving long-context reasoning has become increasingly important due to the rapid emergence of agent applications~\citep{zhao-etal-2024-longagent,  kimiteam2025kimik2openagentic, miromind2025mirothinker, prabhakar2025enterprisedeepresearch, gandhi2026endlessterminalsscalingrl}. 
% Thus, developing scalable and principled training methods that move beyond synthetic context expansion and provide effective supervision for long-context training is becoming urgent for advancing large language models.
\section{Conclusion}

In this work, we present an unsupervised reinforcement learning framework for improving the long-context capabilities of LLMs. 
By formulating reconstruction as a sequential decision-making problem with verifiable rewards derived directly from raw documents, our approach eliminates the need for manually curated long-context data or teacher-model supervision. 
This enables a scalable and principled alternative to existing long-context training paradigms.
Extensive experiments on RULER and LongBench v2 demonstrate that reconstruction-based RLVR effectively enhances long-context performance across multiple backbone models and context lengths. 
Beyond empirical gains, our findings highlight document structure itself as a valuable and underexplored supervision signal. 
We hope that this work motivates further research into unsupervised and self-supervised training objectives that leverage intrinsic structure in raw data and contributes to the development of more capable and scalable long-context language models.
\section*{Limitations}

Despite its effectiveness, our approach has several limitations. 
Our method requires access to sufficiently long and well-structured documents, which may limit its applicability in domains where long-form data is scarce or noisy. 
In addition, we observe that the benefits of reconstruction training vary across backbone models.
And models may have different requirements about document quality and length.
Finally, while our method scales well within the evaluated data range, its behavior at substantially larger scales and with different model sizes remains an open area for future work.

\bibliography{anthology,custom}

@misc{yu2025dapoopensourcellmreinforcement,
      title={DAPO: An Open-Source LLM Reinforcement Learning System at Scale}, 
      author={Qiying Yu and Zheng Zhang and Ruofei Zhu and Yufeng Yuan and Xiaochen Zuo and Yu Yue and Weinan Dai and Tiantian Fan and Gaohong Liu and Lingjun Liu and Xin Liu and Haibin Lin and Zhiqi Lin and Bole Ma and Guangming Sheng and Yuxuan Tong and Chi Zhang and Mofan Zhang and Wang Zhang and Hang Zhu and Jinhua Zhu and Jiaze Chen and Jiangjie Chen and Chengyi Wang and Hongli Yu and Yuxuan Song and Xiangpeng Wei and Hao Zhou and Jingjing Liu and Wei-Ying Ma and Ya-Qin Zhang and Lin Yan and Mu Qiao and Yonghui Wu and Mingxuan Wang},
      year={2025},
      eprint={2503.14476},
      archivePrefix={arXiv},
      primaryClass={cs.LG},
      url={https://arxiv.org/abs/2503.14476}, 
}

@inproceedings{
yeo2025demystifying,
title={Demystifying Long Chain-of-Thought Reasoning in {LLM}s},
author={Edward Yeo and Yuxuan Tong and Xinyao Niu and Graham Neubig and Xiang Yue},
booktitle={ICLR 2025 Workshop on Navigating and Addressing Data Problems for Foundation Models},
year={2025},
url={https://openreview.net/forum?id=AgtQlhMQ0V}
}

@inproceedings{
zeng2025simplerlzoo,
title={Simple{RL}-Zoo: Investigating and Taming Zero Reinforcement Learning for Open Base Models in the Wild},
author={Weihao Zeng and Yuzhen Huang and Qian Liu and Wei Liu and Keqing He and Zejun MA and Junxian He},
booktitle={Second Conference on Language Modeling},
year={2025},
url={https://openreview.net/forum?id=vSMCBUgrQj}
}

@inproceedings{
liu2025prorl,
title={Pro{RL}: Prolonged Reinforcement Learning Expands Reasoning Boundaries in Large Language Models},
author={Mingjie Liu and Shizhe Diao and Ximing Lu and Jian Hu and Xin Dong and Yejin Choi and Jan Kautz and Yi Dong},
booktitle={The Thirty-ninth Annual Conference on Neural Information Processing Systems},
year={2025},
url={https://openreview.net/forum?id=YPsJha5HXQ}
}

@article{Guo_2025,
   title={DeepSeek-R1 incentivizes reasoning in LLMs through reinforcement learning},
   volume={645},
   ISSN={1476-4687},
   url={http://dx.doi.org/10.1038/s41586-025-09422-z},
   DOI={10.1038/s41586-025-09422-z},
   number={8081},
   journal={Nature},
   publisher={Springer Science and Business Media LLC},
   author={Guo, Daya and Yang, Dejian and Zhang, Haowei and Song, Junxiao and Wang, Peiyi and Zhu, Qihao and Xu, Runxin and Zhang, Ruoyu and Ma, Shirong and Bi, Xiao and Zhang, Xiaokang and Yu, Xingkai and Wu, Yu and Wu, Z. F. and Gou, Zhibin and Shao, Zhihong and Li, Zhuoshu and Gao, Ziyi and Liu, Aixin and Xue, Bing and Wang, Bingxuan and Wu, Bochao and Feng, Bei and Lu, Chengda and Zhao, Chenggang and Deng, Chengqi and Ruan, Chong and Dai, Damai and Chen, Deli and Ji, Dongjie and Li, Erhang and Lin, Fangyun and Dai, Fucong and Luo, Fuli and Hao, Guangbo and Chen, Guanting and Li, Guowei and Zhang, H. and Xu, Hanwei and Ding, Honghui and Gao, Huazuo and Qu, Hui and Li, Hui and Guo, Jianzhong and Li, Jiashi and Chen, Jingchang and Yuan, Jingyang and Tu, Jinhao and Qiu, Junjie and Li, Junlong and Cai, J. L. and Ni, Jiaqi and Liang, Jian and Chen, Jin and Dong, Kai and Hu, Kai and You, Kaichao and Gao, Kaige and Guan, Kang and Huang, Kexin and Yu, Kuai and Wang, Lean and Zhang, Lecong and Zhao, Liang and Wang, Litong and Zhang, Liyue and Xu, Lei and Xia, Leyi and Zhang, Mingchuan and Zhang, Minghua and Tang, Minghui and Zhou, Mingxu and Li, Meng and Wang, Miaojun and Li, Mingming and Tian, Ning and Huang, Panpan and Zhang, Peng and Wang, Qiancheng and Chen, Qinyu and Du, Qiushi and Ge, Ruiqi and Zhang, Ruisong and Pan, Ruizhe and Wang, Runji and Chen, R. J. and Jin, R. L. and Chen, Ruyi and Lu, Shanghao and Zhou, Shangyan and Chen, Shanhuang and Ye, Shengfeng and Wang, Shiyu and Yu, Shuiping and Zhou, Shunfeng and Pan, Shuting and Li, S. S. and Zhou, Shuang and Wu, Shaoqing and Yun, Tao and Pei, Tian and Sun, Tianyu and Wang, T. and Zeng, Wangding and Liu, Wen and Liang, Wenfeng and Gao, Wenjun and Yu, Wenqin and Zhang, Wentao and Xiao, W. L. and An, Wei and Liu, Xiaodong and Wang, Xiaohan and Chen, Xiaokang and Nie, Xiaotao and Cheng, Xin and Liu, Xin and Xie, Xin and Liu, Xingchao and Yang, Xinyu and Li, Xinyuan and Su, Xuecheng and Lin, Xuheng and Li, X. Q. and Jin, Xiangyue and Shen, Xiaojin and Chen, Xiaosha and Sun, Xiaowen and Wang, Xiaoxiang and Song, Xinnan and Zhou, Xinyi and Wang, Xianzu and Shan, Xinxia and Li, Y. K. and Wang, Y. Q. and Wei, Y. X. and Zhang, Yang and Xu, Yanhong and Li, Yao and Zhao, Yao and Sun, Yaofeng and Wang, Yaohui and Yu, Yi and Zhang, Yichao and Shi, Yifan and Xiong, Yiliang and He, Ying and Piao, Yishi and Wang, Yisong and Tan, Yixuan and Ma, Yiyang and Liu, Yiyuan and Guo, Yongqiang and Ou, Yuan and Wang, Yuduan and Gong, Yue and Zou, Yuheng and He, Yujia and Xiong, Yunfan and Luo, Yuxiang and You, Yuxiang and Liu, Yuxuan and Zhou, Yuyang and Zhu, Y. X. and Huang, Yanping and Li, Yaohui and Zheng, Yi and Zhu, Yuchen and Ma, Yunxian and Tang, Ying and Zha, Yukun and Yan, Yuting and Ren, Z. Z. and Ren, Zehui and Sha, Zhangli and Fu, Zhe and Xu, Zhean and Xie, Zhenda and Zhang, Zhengyan and Hao, Zhewen and Ma, Zhicheng and Yan, Zhigang and Wu, Zhiyu and Gu, Zihui and Zhu, Zijia and Liu, Zijun and Li, Zilin and Xie, Ziwei and Song, Ziyang and Pan, Zizheng and Huang, Zhen and Xu, Zhipeng and Zhang, Zhongyu and Zhang, Zhen},
   year={2025},
   month=sep, pages={633–638} 
}

@misc{yang2025qwen3technicalreport,
      title={Qwen3 Technical Report}, 
      author={An Yang and Anfeng Li and Baosong Yang and Beichen Zhang and Binyuan Hui and Bo Zheng and Bowen Yu and Chang Gao and Chengen Huang and Chenxu Lv and Chujie Zheng and Dayiheng Liu and Fan Zhou and Fei Huang and Feng Hu and Hao Ge and Haoran Wei and Huan Lin and Jialong Tang and Jian Yang and Jianhong Tu and Jianwei Zhang and Jianxin Yang and Jiaxi Yang and Jing Zhou and Jingren Zhou and Junyang Lin and Kai Dang and Keqin Bao and Kexin Yang and Le Yu and Lianghao Deng and Mei Li and Mingfeng Xue and Mingze Li and Pei Zhang and Peng Wang and Qin Zhu and Rui Men and Ruize Gao and Shixuan Liu and Shuang Luo and Tianhao Li and Tianyi Tang and Wenbiao Yin and Xingzhang Ren and Xinyu Wang and Xinyu Zhang and Xuancheng Ren and Yang Fan and Yang Su and Yichang Zhang and Yinger Zhang and Yu Wan and Yuqiong Liu and Zekun Wang and Zeyu Cui and Zhenru Zhang and Zhipeng Zhou and Zihan Qiu},
      year={2025},
      eprint={2505.09388},
      archivePrefix={arXiv},
      primaryClass={cs.CL},
      url={https://arxiv.org/abs/2505.09388}, 
}

@article{miromind2025mirothinker,
  title={MiroThinker: Pushing the Performance Boundaries of Open-Source Research Agents via Model, Context, and Interactive Scaling},
  author={MiroMind Team and Bai, Song and Bing, Lidong and Chen, Carson and Chen, Guanzheng and Chen, Yuntao and Chen, Zhe and Chen, Ziyi and Dai, Jifeng and Dong, Xuan and others},
  journal={arXiv preprint arXiv:2511.11793},
  year={2025}
}

@misc{tongyideepresearchteam2025tongyideepresearchtechnicalreport,
      title={Tongyi DeepResearch Technical Report}, 
      author={Tongyi DeepResearch Team and Baixuan Li and Bo Zhang and Dingchu Zhang and Fei Huang and Guangyu Li and Guoxin Chen and Huifeng Yin and Jialong Wu and Jingren Zhou and Kuan Li and Liangcai Su and Litu Ou and Liwen Zhang and Pengjun Xie and Rui Ye and Wenbiao Yin and Xinmiao Yu and Xinyu Wang and Xixi Wu and Xuanzhong Chen and Yida Zhao and Zhen Zhang and Zhengwei Tao and Zhongwang Zhang and Zile Qiao and Chenxi Wang and Donglei Yu and Gang Fu and Haiyang Shen and Jiayin Yang and Jun Lin and Junkai Zhang and Kui Zeng and Li Yang and Hailong Yin and Maojia Song and Ming Yan and Minpeng Liao and Peng Xia and Qian Xiao and Rui Min and Ruixue Ding and Runnan Fang and Shaowei Chen and Shen Huang and Shihang Wang and Shihao Cai and Weizhou Shen and Xiaobin Wang and Xin Guan and Xinyu Geng and Yingcheng Shi and Yuning Wu and Zhuo Chen and Zijian Li and Yong Jiang},
      year={2025},
      eprint={2510.24701},
      archivePrefix={arXiv},
      primaryClass={cs.CL},
      url={https://arxiv.org/abs/2510.24701}, 
}

@article{prabhakar2025enterprisedeepresearch,
  title={Enterprise Deep Research: Steerable Multi-Agent Deep Research for Enterprise Analytics},
  author={Prabhakar, Akshara and Ram, Roshan and Chen, Zixiang and Savarese, Silvio and Wang, Frank and Xiong, Caiming and Wang, Huan and Yao, Weiran},
  journal={arXiv preprint arXiv:2510.17797},
  year={2025}
}

@misc{wu2025longwriterzeromasteringultralongtext,
      title={LongWriter-Zero: Mastering Ultra-Long Text Generation via Reinforcement Learning}, 
      author={Yuhao Wu and Yushi Bai and Zhiqiang Hu and Roy Ka-Wei Lee and Juanzi Li},
      year={2025},
      eprint={2506.18841},
      archivePrefix={arXiv},
      primaryClass={cs.CL},
      url={https://arxiv.org/abs/2506.18841}, 
}

@inproceedings{
zhuang2025scaling,
title={Scaling Long Context Training Data by Long-Distance Referrals},
author={Yonghao Zhuang and Lanxiang Hu and Longfei Yun and Souvik Kundu and Zhengzhong Liu and Eric P. Xing and Hao Zhang},
booktitle={The Thirteenth International Conference on Learning Representations},
year={2025},
url={https://openreview.net/forum?id=tePFpDgyqg}
}

@inproceedings{
wu2025retrieval,
title={Retrieval Head Mechanistically Explains Long-Context Factuality},
author={Wenhao Wu and Yizhong Wang and Guangxuan Xiao and Hao Peng and Yao Fu},
booktitle={The Thirteenth International Conference on Learning Representations},
year={2025},
url={https://openreview.net/forum?id=EytBpUGB1Z}
}

@inproceedings{
bai2025longwriter,
title={LongWriter: Unleashing 10,000+ Word Generation from Long Context {LLM}s},
author={Yushi Bai and Jiajie Zhang and Xin Lv and Linzhi Zheng and Siqi Zhu and Lei Hou and Yuxiao Dong and Jie Tang and Juanzi Li},
booktitle={The Thirteenth International Conference on Learning Representations},
year={2025},
url={https://openreview.net/forum?id=kQ5s9Yh0WI}
}

@inproceedings{gao-etal-2025-train,
    title = "How to Train Long-Context Language Models (Effectively)",
    author = "Gao, Tianyu  and
      Wettig, Alexander  and
      Yen, Howard  and
      Chen, Danqi",
    editor = "Che, Wanxiang  and
      Nabende, Joyce  and
      Shutova, Ekaterina  and
      Pilehvar, Mohammad Taher",
    booktitle = "Proceedings of the 63rd Annual Meeting of the Association for Computational Linguistics (Volume 1: Long Papers)",
    month = jul,
    year = "2025",
    address = "Vienna, Austria",
    publisher = "Association for Computational Linguistics",
    url = "https://aclanthology.org/2025.acl-long.366/",
    doi = "10.18653/v1/2025.acl-long.366",
    pages = "7376--7399",
    ISBN = "979-8-89176-251-0"
}

@inproceedings{
peng2024yarn,
title={Ya{RN}: Efficient Context Window Extension of Large Language Models},
author={Bowen Peng and Jeffrey Quesnelle and Honglu Fan and Enrico Shippole},
booktitle={The Twelfth International Conference on Learning Representations},
year={2024},
url={https://openreview.net/forum?id=wHBfxhZu1u}
}

@misc{lu2025mobamixtureblockattention,
      title={MoBA: Mixture of Block Attention for Long-Context LLMs}, 
      author={Enzhe Lu and Zhejun Jiang and Jingyuan Liu and Yulun Du and Tao Jiang and Chao Hong and Shaowei Liu and Weiran He and Enming Yuan and Yuzhi Wang and Zhiqi Huang and Huan Yuan and Suting Xu and Xinran Xu and Guokun Lai and Yanru Chen and Huabin Zheng and Junjie Yan and Jianlin Su and Yuxin Wu and Neo Y. Zhang and Zhilin Yang and Xinyu Zhou and Mingxing Zhang and Jiezhong Qiu},
      year={2025},
      eprint={2502.13189},
      archivePrefix={arXiv},
      primaryClass={cs.LG},
      url={https://arxiv.org/abs/2502.13189}, 
}

@misc{yang2025qwen251mtechnicalreport,
      title={Qwen2.5-1M Technical Report}, 
      author={An Yang and Bowen Yu and Chengyuan Li and Dayiheng Liu and Fei Huang and Haoyan Huang and Jiandong Jiang and Jianhong Tu and Jianwei Zhang and Jingren Zhou and Junyang Lin and Kai Dang and Kexin Yang and Le Yu and Mei Li and Minmin Sun and Qin Zhu and Rui Men and Tao He and Weijia Xu and Wenbiao Yin and Wenyuan Yu and Xiafei Qiu and Xingzhang Ren and Xinlong Yang and Yong Li and Zhiying Xu and Zipeng Zhang},
      year={2025},
      eprint={2501.15383},
      archivePrefix={arXiv},
      primaryClass={cs.CL},
      url={https://arxiv.org/abs/2501.15383}, 
}

@article{liu-etal-2024-lost,
    title = "Lost in the Middle: How Language Models Use Long Contexts",
    author = "Liu, Nelson F.  and
      Lin, Kevin  and
      Hewitt, John  and
      Paranjape, Ashwin  and
      Bevilacqua, Michele  and
      Petroni, Fabio  and
      Liang, Percy",
    journal = "Transactions of the Association for Computational Linguistics",
    volume = "12",
    year = "2024",
    address = "Cambridge, MA",
    publisher = "MIT Press",
    url = "https://aclanthology.org/2024.tacl-1.9/",
    doi = "10.1162/tacl_a_00638",
    pages = "157--173"
}

@inproceedings{hsieh-etal-2024-found,
    title = "Found in the middle: Calibrating Positional Attention Bias Improves Long Context Utilization",
    author = "Hsieh, Cheng-Yu  and
      Chuang, Yung-Sung  and
      Li, Chun-Liang  and
      Wang, Zifeng  and
      Le, Long  and
      Kumar, Abhishek  and
      Glass, James  and
      Ratner, Alexander  and
      Lee, Chen-Yu  and
      Krishna, Ranjay  and
      Pfister, Tomas",
    editor = "Ku, Lun-Wei  and
      Martins, Andre  and
      Srikumar, Vivek",
    booktitle = "Findings of the Association for Computational Linguistics: ACL 2024",
    month = aug,
    year = "2024",
    address = "Bangkok, Thailand",
    publisher = "Association for Computational Linguistics",
    url = "https://aclanthology.org/2024.findings-acl.890/",
    doi = "10.18653/v1/2024.findings-acl.890",
    pages = "14982--14995"
}

@inproceedings{
hsieh2024ruler,
title={{RULER}: What{\textquoteright}s the Real Context Size of Your Long-Context Language Models?},
author={Cheng-Ping Hsieh and Simeng Sun and Samuel Kriman and Shantanu Acharya and Dima Rekesh and Fei Jia and Boris Ginsburg},
booktitle={First Conference on Language Modeling},
year={2024},
url={https://openreview.net/forum?id=kIoBbc76Sy}
}

@inproceedings{du-etal-2025-context,
    title = "Context Length Alone Hurts {LLM} Performance Despite Perfect Retrieval",
    author = "Du, Yufeng  and
      Tian, Minyang  and
      Ronanki, Srikanth  and
      Rongali, Subendhu  and
      Bodapati, Sravan Babu  and
      Galstyan, Aram  and
      Wells, Azton  and
      Schwartz, Roy  and
      Huerta, Eliu A  and
      Peng, Hao",
    editor = "Christodoulopoulos, Christos  and
      Chakraborty, Tanmoy  and
      Rose, Carolyn  and
      Peng, Violet",
    booktitle = "Findings of the Association for Computational Linguistics: EMNLP 2025",
    month = nov,
    year = "2025",
    address = "Suzhou, China",
    publisher = "Association for Computational Linguistics",
    url = "https://aclanthology.org/2025.findings-emnlp.1264/",
    doi = "10.18653/v1/2025.findings-emnlp.1264",
    pages = "23281--23298",
    ISBN = "979-8-89176-335-7"
}

@misc{wang2025loongrlreinforcementlearningadvanced,
      title={LoongRL: Reinforcement Learning for Advanced Reasoning over Long Contexts}, 
      author={Siyuan Wang and Gaokai Zhang and Li Lyna Zhang and Ning Shang and Fan Yang and Dongyao Chen and Mao Yang},
      year={2025},
      eprint={2510.19363},
      archivePrefix={arXiv},
      primaryClass={cs.CL},
      url={https://arxiv.org/abs/2510.19363}, 
}

@inproceedings{
chen2025longpo,
title={Long{PO}: Long Context Self-Evolution of Large Language Models through Short-to-Long Preference Optimization},
author={Guanzheng Chen and Xin Li and Michael Shieh and Lidong Bing},
booktitle={The Thirteenth International Conference on Learning Representations},
year={2025},
url={https://openreview.net/forum?id=qTrEq31Shm}
}

@inproceedings{zhang-etal-2025-longreward,
    title = "{L}ong{R}eward: Improving Long-context Large Language Models with {AI} Feedback",
    author = "Zhang, Jiajie  and
      Hou, Zhongni  and
      Lv, Xin  and
      Cao, Shulin  and
      Hou, Zhenyu  and
      Niu, Yilin  and
      Hou, Lei  and
      Dong, Yuxiao  and
      Feng, Ling  and
      Li, Juanzi",
    editor = "Che, Wanxiang  and
      Nabende, Joyce  and
      Shutova, Ekaterina  and
      Pilehvar, Mohammad Taher",
    booktitle = "Proceedings of the 63rd Annual Meeting of the Association for Computational Linguistics (Volume 1: Long Papers)",
    month = jul,
    year = "2025",
    address = "Vienna, Austria",
    publisher = "Association for Computational Linguistics",
    url = "https://aclanthology.org/2025.acl-long.187/",
    doi = "10.18653/v1/2025.acl-long.187",
    pages = "3718--3739",
    ISBN = "979-8-89176-251-0"
}

@misc{huang2025reinforcementlearningrubricanchors,
      title={Reinforcement Learning with Rubric Anchors}, 
      author={Zenan Huang and Yihong Zhuang and Guoshan Lu and Zeyu Qin and Haokai Xu and Tianyu Zhao and Ru Peng and Jiaqi Hu and Zhanming Shen and Xiaomeng Hu and Xijun Gu and Peiyi Tu and Jiaxin Liu and Wenyu Chen and Yuzhuo Fu and Zhiting Fan and Yanmei Gu and Yuanyuan Wang and Zhengkai Yang and Jianguo Li and Junbo Zhao},
      year={2025},
      eprint={2508.12790},
      archivePrefix={arXiv},
      primaryClass={cs.AI},
      url={https://arxiv.org/abs/2508.12790}, 
}

@inproceedings{
kim2025rlvr,
title={{RLVR} vs. Distillation: Understanding Accuracy and Capability in {LLM} Mathematical Reasoning},
author={Minwu Kim and Anubhav Shrestha and Safal Shrestha and Aadim Nepal and Keith W. Ross},
booktitle={The 5th Workshop on Mathematical Reasoning and AI at NeurIPS 2025},
year={2025},
url={https://openreview.net/forum?id=DH9hjro5eu}
}

@inproceedings{
cheng2025do,
title={Do Students Debias Like Teachers? On the Distillability of Bias Mitigation Methods},
author={Jiali Cheng and Hadi Amiri},
booktitle={ICML 2025 Workshop on Reliable and Responsible Foundation Models},
year={2025},
url={https://openreview.net/forum?id=LLnjLFeKBJ}
}

@inproceedings{
an2024trainingfree,
title={Training-Free Long-Context Scaling of Large Language Models},
author={Chenxin An and Fei Huang and Jun Zhang and Shansan Gong and Xipeng Qiu and Chang Zhou and Lingpeng Kong},
booktitle={Forty-first International Conference on Machine Learning},
year={2024},
url={https://openreview.net/forum?id=If4xW9vF7U}
}

@inproceedings{
zhu2024pose,
title={Po{SE}: Efficient Context Window Extension of {LLM}s via Positional Skip-wise Training},
author={Dawei Zhu and Nan Yang and Liang Wang and Yifan Song and Wenhao Wu and Furu Wei and Sujian Li},
booktitle={The Twelfth International Conference on Learning Representations},
year={2024},
url={https://openreview.net/forum?id=3Z1gxuAQrA}
}

@inproceedings{bai-etal-2025-longbench,
    title = "{L}ong{B}ench v2: Towards Deeper Understanding and Reasoning on Realistic Long-context Multitasks",
    author = "Bai, Yushi  and
      Tu, Shangqing  and
      Zhang, Jiajie  and
      Peng, Hao  and
      Wang, Xiaozhi  and
      Lv, Xin  and
      Cao, Shulin  and
      Xu, Jiazheng  and
      Hou, Lei  and
      Dong, Yuxiao  and
      Tang, Jie  and
      Li, Juanzi",
    editor = "Che, Wanxiang  and
      Nabende, Joyce  and
      Shutova, Ekaterina  and
      Pilehvar, Mohammad Taher",
    booktitle = "Proceedings of the 63rd Annual Meeting of the Association for Computational Linguistics (Volume 1: Long Papers)",
    month = jul,
    year = "2025",
    address = "Vienna, Austria",
    publisher = "Association for Computational Linguistics",
    url = "https://aclanthology.org/2025.acl-long.183/",
    doi = "10.18653/v1/2025.acl-long.183",
    pages = "3639--3664",
    ISBN = "979-8-89176-251-0"
}

@misc{wu2025visualjigsawposttrainingimproves,
      title={Visual Jigsaw Post-Training Improves MLLMs}, 
      author={Penghao Wu and Yushan Zhang and Haiwen Diao and Bo Li and Lewei Lu and Ziwei Liu},
      year={2025},
      eprint={2509.25190},
      archivePrefix={arXiv},
      primaryClass={cs.CV},
      url={https://arxiv.org/abs/2509.25190}, 
}

@inproceedings{Bengio2009CurriculumL,
  title={Curriculum learning},
  author={Yoshua Bengio and J{\'e}r{\^o}me Louradour and Ronan Collobert and Jason Weston},
  booktitle={International Conference on Machine Learning},
  year={2009},
  url={https://api.semanticscholar.org/CorpusID:873046}
}

@article{sheng2024hybridflow,
  title   = {HybridFlow: A Flexible and Efficient RLHF Framework},
  author  = {Guangming Sheng and Chi Zhang and Zilingfeng Ye and Xibin Wu and Wang Zhang and Ru Zhang and Yanghua Peng and Haibin Lin and Chuan Wu},
  year    = {2024},
  journal = {arXiv preprint arXiv: 2409.19256}
}

@misc{shao2024deepseekmathpushinglimitsmathematical,
      title={DeepSeekMath: Pushing the Limits of Mathematical Reasoning in Open Language Models}, 
      author={Zhihong Shao and Peiyi Wang and Qihao Zhu and Runxin Xu and Junxiao Song and Xiao Bi and Haowei Zhang and Mingchuan Zhang and Y. K. Li and Y. Wu and Daya Guo},
      year={2024},
      eprint={2402.03300},
      archivePrefix={arXiv},
      primaryClass={cs.CL},
      url={https://arxiv.org/abs/2402.03300}, 
}

@misc{lambert2025tulu3pushingfrontiers,
      title={Tulu 3: Pushing Frontiers in Open Language Model Post-Training}, 
      author={Nathan Lambert and Jacob Morrison and Valentina Pyatkin and Shengyi Huang and Hamish Ivison and Faeze Brahman and Lester James V. Miranda and Alisa Liu and Nouha Dziri and Shane Lyu and Yuling Gu and Saumya Malik and Victoria Graf and Jena D. Hwang and Jiangjiang Yang and Ronan Le Bras and Oyvind Tafjord and Chris Wilhelm and Luca Soldaini and Noah A. Smith and Yizhong Wang and Pradeep Dasigi and Hannaneh Hajishirzi},
      year={2025},
      eprint={2411.15124},
      archivePrefix={arXiv},
      primaryClass={cs.CL},
      url={https://arxiv.org/abs/2411.15124}, 
}

@misc{openai2024openaio1card,
      title={OpenAI o1 System Card}, 
      author={OpenAI and : and Aaron Jaech and Adam Kalai and Adam Lerer and Adam Richardson and Ahmed El-Kishky and Aiden Low and Alec Helyar and Aleksander Madry and Alex Beutel and Alex Carney and Alex Iftimie and Alex Karpenko and Alex Tachard Passos and Alexander Neitz and Alexander Prokofiev and Alexander Wei and Allison Tam and Ally Bennett and Ananya Kumar and Andre Saraiva and Andrea Vallone and Andrew Duberstein and Andrew Kondrich and Andrey Mishchenko and Andy Applebaum and Angela Jiang and Ashvin Nair and Barret Zoph and Behrooz Ghorbani and Ben Rossen and Benjamin Sokolowsky and Boaz Barak and Bob McGrew and Borys Minaiev and Botao Hao and Bowen Baker and Brandon Houghton and Brandon McKinzie and Brydon Eastman and Camillo Lugaresi and Cary Bassin and Cary Hudson and Chak Ming Li and Charles de Bourcy and Chelsea Voss and Chen Shen and Chong Zhang and Chris Koch and Chris Orsinger and Christopher Hesse and Claudia Fischer and Clive Chan and Dan Roberts and Daniel Kappler and Daniel Levy and Daniel Selsam and David Dohan and David Farhi and David Mely and David Robinson and Dimitris Tsipras and Doug Li and Dragos Oprica and Eben Freeman and Eddie Zhang and Edmund Wong and Elizabeth Proehl and Enoch Cheung and Eric Mitchell and Eric Wallace and Erik Ritter and Evan Mays and Fan Wang and Felipe Petroski Such and Filippo Raso and Florencia Leoni and Foivos Tsimpourlas and Francis Song and Fred von Lohmann and Freddie Sulit and Geoff Salmon and Giambattista Parascandolo and Gildas Chabot and Grace Zhao and Greg Brockman and Guillaume Leclerc and Hadi Salman and Haiming Bao and Hao Sheng and Hart Andrin and Hessam Bagherinezhad and Hongyu Ren and Hunter Lightman and Hyung Won Chung and Ian Kivlichan and Ian O'Connell and Ian Osband and Ignasi Clavera Gilaberte and Ilge Akkaya and Ilya Kostrikov and Ilya Sutskever and Irina Kofman and Jakub Pachocki and James Lennon and Jason Wei and Jean Harb and Jerry Twore and Jiacheng Feng and Jiahui Yu and Jiayi Weng and Jie Tang and Jieqi Yu and Joaquin Quiñonero Candela and Joe Palermo and Joel Parish and Johannes Heidecke and John Hallman and John Rizzo and Jonathan Gordon and Jonathan Uesato and Jonathan Ward and Joost Huizinga and Julie Wang and Kai Chen and Kai Xiao and Karan Singhal and Karina Nguyen and Karl Cobbe and Katy Shi and Kayla Wood and Kendra Rimbach and Keren Gu-Lemberg and Kevin Liu and Kevin Lu and Kevin Stone and Kevin Yu and Lama Ahmad and Lauren Yang and Leo Liu and Leon Maksin and Leyton Ho and Liam Fedus and Lilian Weng and Linden Li and Lindsay McCallum and Lindsey Held and Lorenz Kuhn and Lukas Kondraciuk and Lukasz Kaiser and Luke Metz and Madelaine Boyd and Maja Trebacz and Manas Joglekar and Mark Chen and Marko Tintor and Mason Meyer and Matt Jones and Matt Kaufer and Max Schwarzer and Meghan Shah and Mehmet Yatbaz and Melody Y. Guan and Mengyuan Xu and Mengyuan Yan and Mia Glaese and Mianna Chen and Michael Lampe and Michael Malek and Michele Wang and Michelle Fradin and Mike McClay and Mikhail Pavlov and Miles Wang and Mingxuan Wang and Mira Murati and Mo Bavarian and Mostafa Rohaninejad and Nat McAleese and Neil Chowdhury and Neil Chowdhury and Nick Ryder and Nikolas Tezak and Noam Brown and Ofir Nachum and Oleg Boiko and Oleg Murk and Olivia Watkins and Patrick Chao and Paul Ashbourne and Pavel Izmailov and Peter Zhokhov and Rachel Dias and Rahul Arora and Randall Lin and Rapha Gontijo Lopes and Raz Gaon and Reah Miyara and Reimar Leike and Renny Hwang and Rhythm Garg and Robin Brown and Roshan James and Rui Shu and Ryan Cheu and Ryan Greene and Saachi Jain and Sam Altman and Sam Toizer and Sam Toyer and Samuel Miserendino and Sandhini Agarwal and Santiago Hernandez and Sasha Baker and Scott McKinney and Scottie Yan and Shengjia Zhao and Shengli Hu and Shibani Santurkar and Shraman Ray Chaudhuri and Shuyuan Zhang and Siyuan Fu and Spencer Papay and Steph Lin and Suchir Balaji and Suvansh Sanjeev and Szymon Sidor and Tal Broda and Aidan Clark and Tao Wang and Taylor Gordon and Ted Sanders and Tejal Patwardhan and Thibault Sottiaux and Thomas Degry and Thomas Dimson and Tianhao Zheng and Timur Garipov and Tom Stasi and Trapit Bansal and Trevor Creech and Troy Peterson and Tyna Eloundou and Valerie Qi and Vineet Kosaraju and Vinnie Monaco and Vitchyr Pong and Vlad Fomenko and Weiyi Zheng and Wenda Zhou and Wes McCabe and Wojciech Zaremba and Yann Dubois and Yinghai Lu and Yining Chen and Young Cha and Yu Bai and Yuchen He and Yuchen Zhang and Yunyun Wang and Zheng Shao and Zhuohan Li},
      year={2024},
      eprint={2412.16720},
      archivePrefix={arXiv},
      primaryClass={cs.AI},
      url={https://arxiv.org/abs/2412.16720}, 
}

@misc{kimiteam2025kimik2openagentic,
      title={Kimi K2: Open Agentic Intelligence}, 
      author={Kimi Team and Yifan Bai and Yiping Bao and Guanduo Chen and Jiahao Chen and Ningxin Chen and Ruijue Chen and Yanru Chen and Yuankun Chen and Yutian Chen and Zhuofu Chen and Jialei Cui and Hao Ding and Mengnan Dong and Angang Du and Chenzhuang Du and Dikang Du and Yulun Du and Yu Fan and Yichen Feng and Kelin Fu and Bofei Gao and Hongcheng Gao and Peizhong Gao and Tong Gao and Xinran Gu and Longyu Guan and Haiqing Guo and Jianhang Guo and Hao Hu and Xiaoru Hao and Tianhong He and Weiran He and Wenyang He and Chao Hong and Yangyang Hu and Zhenxing Hu and Weixiao Huang and Zhiqi Huang and Zihao Huang and Tao Jiang and Zhejun Jiang and Xinyi Jin and Yongsheng Kang and Guokun Lai and Cheng Li and Fang Li and Haoyang Li and Ming Li and Wentao Li and Yanhao Li and Yiwei Li and Zhaowei Li and Zheming Li and Hongzhan Lin and Xiaohan Lin and Zongyu Lin and Chengyin Liu and Chenyu Liu and Hongzhang Liu and Jingyuan Liu and Junqi Liu and Liang Liu and Shaowei Liu and T. Y. Liu and Tianwei Liu and Weizhou Liu and Yangyang Liu and Yibo Liu and Yiping Liu and Yue Liu and Zhengying Liu and Enzhe Lu and Lijun Lu and Shengling Ma and Xinyu Ma and Yingwei Ma and Shaoguang Mao and Jie Mei and Xin Men and Yibo Miao and Siyuan Pan and Yebo Peng and Ruoyu Qin and Bowen Qu and Zeyu Shang and Lidong Shi and Shengyuan Shi and Feifan Song and Jianlin Su and Zhengyuan Su and Xinjie Sun and Flood Sung and Heyi Tang and Jiawen Tao and Qifeng Teng and Chensi Wang and Dinglu Wang and Feng Wang and Haiming Wang and Jianzhou Wang and Jiaxing Wang and Jinhong Wang and Shengjie Wang and Shuyi Wang and Yao Wang and Yejie Wang and Yiqin Wang and Yuxin Wang and Yuzhi Wang and Zhaoji Wang and Zhengtao Wang and Zhexu Wang and Chu Wei and Qianqian Wei and Wenhao Wu and Xingzhe Wu and Yuxin Wu and Chenjun Xiao and Xiaotong Xie and Weimin Xiong and Boyu Xu and Jing Xu and Jinjing Xu and L. H. Xu and Lin Xu and Suting Xu and Weixin Xu and Xinran Xu and Yangchuan Xu and Ziyao Xu and Junjie Yan and Yuzi Yan and Xiaofei Yang and Ying Yang and Zhen Yang and Zhilin Yang and Zonghan Yang and Haotian Yao and Xingcheng Yao and Wenjie Ye and Zhuorui Ye and Bohong Yin and Longhui Yu and Enming Yuan and Hongbang Yuan and Mengjie Yuan and Haobing Zhan and Dehao Zhang and Hao Zhang and Wanlu Zhang and Xiaobin Zhang and Yangkun Zhang and Yizhi Zhang and Yongting Zhang and Yu Zhang and Yutao Zhang and Yutong Zhang and Zheng Zhang and Haotian Zhao and Yikai Zhao and Huabin Zheng and Shaojie Zheng and Jianren Zhou and Xinyu Zhou and Zaida Zhou and Zhen Zhu and Weiyu Zhuang and Xinxing Zu},
      year={2025},
      eprint={2507.20534},
      archivePrefix={arXiv},
      primaryClass={cs.LG},
      url={https://arxiv.org/abs/2507.20534}, 
}

@misc{huang2025winninggoldimo2025,
      title={Winning Gold at IMO 2025 with a Model-Agnostic Verification-and-Refinement Pipeline}, 
      author={Yichen Huang and Lin F. Yang},
      year={2025},
      eprint={2507.15855},
      archivePrefix={arXiv},
      primaryClass={cs.AI},
      url={https://arxiv.org/abs/2507.15855}, 
}

@misc{yue2025doesreinforcementlearningreally,
      title={Does Reinforcement Learning Really Incentivize Reasoning Capacity in LLMs Beyond the Base Model?}, 
      author={Yang Yue and Zhiqi Chen and Rui Lu and Andrew Zhao and Zhaokai Wang and Yang Yue and Shiji Song and Gao Huang},
      year={2025},
      eprint={2504.13837},
      archivePrefix={arXiv},
      primaryClass={cs.AI},
      url={https://arxiv.org/abs/2504.13837}, 
}

@misc{gandhi2025cognitivebehaviorsenableselfimproving,
      title={Cognitive Behaviors that Enable Self-Improving Reasoners, or, Four Habits of Highly Effective STaRs}, 
      author={Kanishk Gandhi and Ayush Chakravarthy and Anikait Singh and Nathan Lile and Noah D. Goodman},
      year={2025},
      eprint={2503.01307},
      archivePrefix={arXiv},
      primaryClass={cs.CL},
      url={https://arxiv.org/abs/2503.01307}, 
}

@misc{wan2025qwenlongl1longcontextlargereasoning,
      title={QwenLong-L1: Towards Long-Context Large Reasoning Models with Reinforcement Learning}, 
      author={Fanqi Wan and Weizhou Shen and Shengyi Liao and Yingcheng Shi and Chenliang Li and Ziyi Yang and Ji Zhang and Fei Huang and Jingren Zhou and Ming Yan},
      year={2025},
      eprint={2505.17667},
      archivePrefix={arXiv},
      primaryClass={cs.CL},
      url={https://arxiv.org/abs/2505.17667}, 
}

@inproceedings{
yen2025helmet,
title={{HELMET}: How to Evaluate Long-context Models Effectively and Thoroughly},
author={Howard Yen and Tianyu Gao and Minmin Hou and Ke Ding and Daniel Fleischer and Peter Izsak and Moshe Wasserblat and Danqi Chen},
booktitle={The Thirteenth International Conference on Learning Representations},
year={2025},
url={https://openreview.net/forum?id=293V3bJbmE}
}

@misc{
chen2025on,
title={On the Diversity of Synthetic Data and its Impact on Training Large Language Models},
author={Hao Chen and Abdul Waheed and Xiang Li and Yidong Wang and Jindong Wang and Bhiksha Raj and Marah I Abdin},
year={2025},
url={https://openreview.net/forum?id=oqsQbn4XfT}
}

@inproceedings{li-etal-2024-making,
    title = "Making Long-Context Language Models Better Multi-Hop Reasoners",
    author = "Li, Yanyang  and
      Liang, Shuo  and
      Lyu, Michael  and
      Wang, Liwei",
    editor = "Ku, Lun-Wei  and
      Martins, Andre  and
      Srikumar, Vivek",
    booktitle = "Proceedings of the 62nd Annual Meeting of the Association for Computational Linguistics (Volume 1: Long Papers)",
    month = aug,
    year = "2024",
    address = "Bangkok, Thailand",
    publisher = "Association for Computational Linguistics",
    url = "https://aclanthology.org/2024.acl-long.135/",
    doi = "10.18653/v1/2024.acl-long.135",
    pages = "2462--2475"
}

@inproceedings{zhao-etal-2024-longagent,
    title = "{LONGAGENT}: Achieving Question Answering for 128k-Token-Long Documents through Multi-Agent Collaboration",
    author = "Zhao, Jun  and
      Zu, Can  and
      Hao, Xu  and
      Lu, Yi  and
      He, Wei  and
      Ding, Yiwen  and
      Gui, Tao  and
      Zhang, Qi  and
      Huang, Xuanjing",
    editor = "Al-Onaizan, Yaser  and
      Bansal, Mohit  and
      Chen, Yun-Nung",
    booktitle = "Proceedings of the 2024 Conference on Empirical Methods in Natural Language Processing",
    month = nov,
    year = "2024",
    address = "Miami, Florida, USA",
    publisher = "Association for Computational Linguistics",
    url = "https://aclanthology.org/2024.emnlp-main.912/",
    doi = "10.18653/v1/2024.emnlp-main.912",
    pages = "16310--16324"
}

@inproceedings{NEURIPS2019_dc6a7e65,
 author = {Yang, Zhilin and Dai, Zihang and Yang, Yiming and Carbonell, Jaime and Salakhutdinov, Russ R and Le, Quoc V},
 booktitle = {Advances in Neural Information Processing Systems},
 editor = {H. Wallach and H. Larochelle and A. Beygelzimer and F. d\textquotesingle Alch\'{e}-Buc and E. Fox and R. Garnett},
 pages = {},
 publisher = {Curran Associates, Inc.},
 title = {XLNet: Generalized Autoregressive Pretraining for Language Understanding},
 url = {https://proceedings.neurips.cc/paper_files/paper/2019/file/dc6a7e655d7e5840e66733e9ee67cc69-Paper.pdf},
 volume = {32},
 year = {2019}
}

@inproceedings{
kumar2025training,
title={Training Language Models to Self-Correct via Reinforcement Learning},
author={Aviral Kumar and Vincent Zhuang and Rishabh Agarwal and Yi Su and John D Co-Reyes and Avi Singh and Kate Baumli and Shariq Iqbal and Colton Bishop and Rebecca Roelofs and Lei M Zhang and Kay McKinney and Disha Shrivastava and Cosmin Paduraru and George Tucker and Doina Precup and Feryal Behbahani and Aleksandra Faust},
booktitle={The Thirteenth International Conference on Learning Representations},
year={2025},
url={https://openreview.net/forum?id=CjwERcAU7w}
}

@misc{claude_ai,
  author       = {{Anthropic, PBC}},
  title        = {Claude},
  howpublished = {\url{https://www.claude.ai}},
  year         = {2025},
  note         = {Accessed: 2025-10-01}
}

@misc{mo2025multiagenttoolintegratedpolicyoptimization,
      title={Multi-Agent Tool-Integrated Policy Optimization}, 
      author={Zhanfeng Mo and Xingxuan Li and Yuntao Chen and Lidong Bing},
      year={2025},
      eprint={2510.04678},
      archivePrefix={arXiv},
      primaryClass={cs.CL},
      url={https://arxiv.org/abs/2510.04678}, 
}

@misc{gandhi2026endlessterminalsscalingrl,
      title={Endless Terminals: Scaling RL Environments for Terminal Agents}, 
      author={Kanishk Gandhi and Shivam Garg and Noah D. Goodman and Dimitris Papailiopoulos},
      year={2026},
      eprint={2601.16443},
      archivePrefix={arXiv},
      primaryClass={cs.LG},
      url={https://arxiv.org/abs/2601.16443}, 
}

@misc{wang2026visgymdiversecustomizablescalable,
      title={VisGym: Diverse, Customizable, Scalable Environments for Multimodal Agents}, 
      author={Zirui Wang and Junyi Zhang and Jiaxin Ge and Long Lian and Letian Fu and Lisa Dunlap and Ken Goldberg and XuDong Wang and Ion Stoica and David M. Chan and Sewon Min and Joseph E. Gonzalez},
      year={2026},
      eprint={2601.16973},
      archivePrefix={arXiv},
      primaryClass={cs.CV},
      url={https://arxiv.org/abs/2601.16973}, 
}

@misc{lu2026goldengoosesimpletrick,
      title={Golden Goose: A Simple Trick to Synthesize Unlimited RLVR Tasks from Unverifiable Internet Text}, 
      author={Ximing Lu and David Acuna and Jaehun Jung and Jian Hu and Di Zhang and Shizhe Diao and Yunheng Zou and Shaokun Zhang and Brandon Cui and Mingjie Liu and Hyunwoo Kim and Prithviraj Ammanabrolu and Jan Kautz and Yi Dong and Yejin Choi},
      year={2026},
      eprint={2601.22975},
      archivePrefix={arXiv},
      primaryClass={cs.AI},
      url={https://arxiv.org/abs/2601.22975}, 
}

@misc{xu2026alternatingreinforcementlearningrubricbased,
      title={Alternating Reinforcement Learning for Rubric-Based Reward Modeling in Non-Verifiable LLM Post-Training}, 
      author={Ran Xu and Tianci Liu and Zihan Dong and Tony You and Ilgee Hong and Carl Yang and Linjun Zhang and Tao Zhao and Haoyu Wang},
      year={2026},
      eprint={2602.01511},
      archivePrefix={arXiv},
      primaryClass={cs.CL},
      url={https://arxiv.org/abs/2602.01511}, 
}

@misc{zeng2025rlvescalingreinforcementlearning,
      title={RLVE: Scaling Up Reinforcement Learning for Language Models with Adaptive Verifiable Environments}, 
      author={Zhiyuan Zeng and Hamish Ivison and Yiping Wang and Lifan Yuan and Shuyue Stella Li and Zhuorui Ye and Siting Li and Jacqueline He and Runlong Zhou and Tong Chen and Chenyang Zhao and Yulia Tsvetkov and Simon Shaolei Du and Natasha Jaques and Hao Peng and Pang Wei Koh and Hannaneh Hajishirzi},
      year={2025},
      eprint={2511.07317},
      archivePrefix={arXiv},
      primaryClass={cs.CL},
      url={https://arxiv.org/abs/2511.07317}, 
}

@misc{wang2026rlanythingforgeenvironmentpolicy,
      title={RLAnything: Forge Environment, Policy, and Reward Model in Completely Dynamic RL System}, 
      author={Yinjie Wang and Tianbao Xie and Ke Shen and Mengdi Wang and Ling Yang},
      year={2026},
      eprint={2602.02488},
      archivePrefix={arXiv},
      primaryClass={cs.LG},
      url={https://arxiv.org/abs/2602.02488}, 
}

@inproceedings{
chen2026longrlvr,
title={Long{RLVR}: Long-Context Reinforcement Learning Requires Verifiable Context Rewards},
author={Guanzheng Chen and Michael Qizhe Shieh and Lidong Bing},
booktitle={The Fourteenth International Conference on Learning Representations},
year={2026},
url={https://openreview.net/forum?id=omVhYvyTPJ}
}

\appendix

\section{Appendix}

\subsection{Reconstruction Prompt}
\label{reprompt}

We append the reconstruction prompt below.

\begin{tcolorbox}[
  enhanced,
  colback=blue!3,
  colframe=black,
  boxrule=1.2pt,
  arc=6pt,
  left=8pt,
  right=8pt,
  top=10pt,
  bottom=8pt,
  attach boxed title to top left={
    xshift=6pt,
    yshift=-3mm
  },
  boxed title style={
    colback=black,
    colframe=black,
    arc=4pt,
    left=6pt,
    right=6pt,
    top=4pt,
    bottom=4pt
  },
  title=\textcolor{white}{\bfseries Reconstruction Prompt}
]

The following document contains missing segments marked as
\texttt{\textless C\_i\textgreater MISSING\texttt{\textless /C\_i\textgreater}}.

Please reason about the logical and narrative structure of the document and select appropriate chunks one by one from the given options to reconstruct it.

Then, output the label for each missing chunk by order in \texttt{\char`\\ boxed\{\}} separated by commas.

The document is as follows: $\{\texttt{corrupted document}\}$

The options are: $\{\texttt{options}\}$

\end{tcolorbox}

\subsection{On continual Pretraining}
\label{continue}

We further explore continual pretraining by training the models for one epoch on our curated long-document corpus. 
However, this approach results in significantly worse performance compared to the original models. 
We attribute this degradation to two main factors. 
First, the quality of our collected long documents may not exceed that of the proprietary data used during the original pretraining of the models. 
Second, continual pretraining on instruction-tuned models may disrupt their instruction-following capabilities. 
As a result, continual pretraining of LLaMA-3.1-8B-Instruct and Qwen2.5-7B-Instruct-1M does not yield performance improvements.

\subsection{On Validation}
\label{valid}

\begin{figure*}[t]
    \centering
    \includegraphics[width=0.8\linewidth]{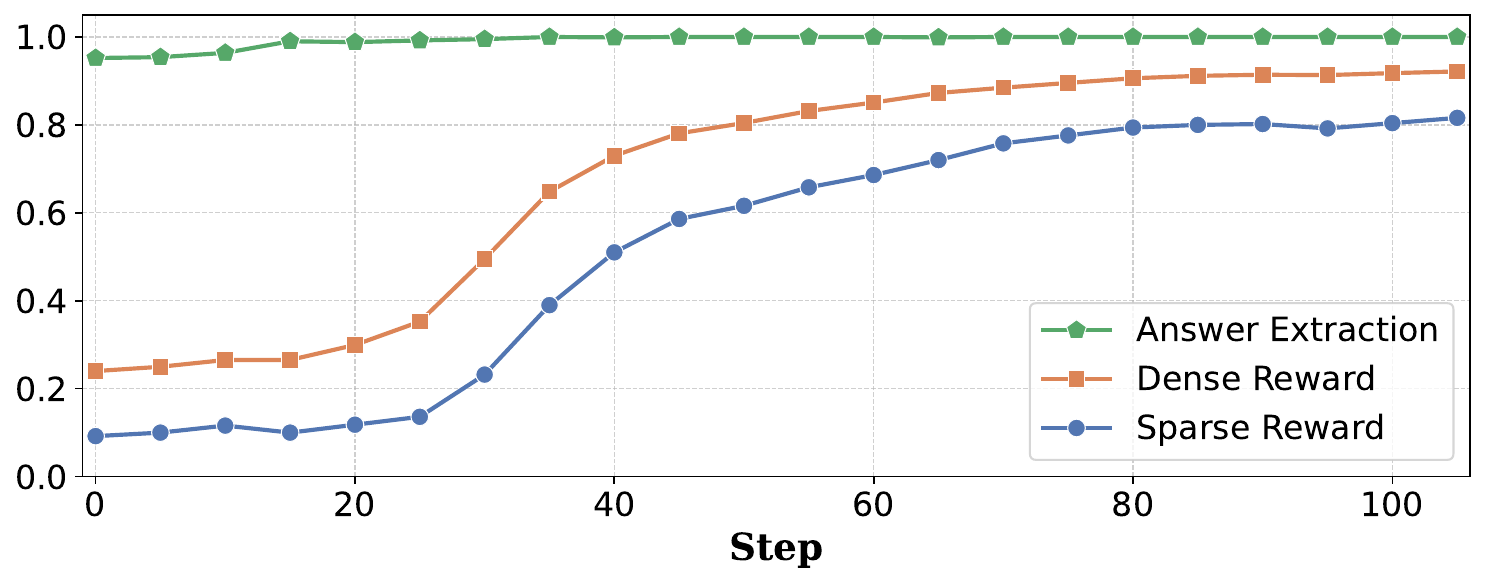}
    \caption{We report metrics on validation set during training process.}
    \label{fig:valid}
\end{figure*}

During training, we monitor model performance on a held-out validation set to assess
optimization stability and learning dynamics in reconstruction training. 
Specifically, we track three metrics: (1) the success rate of answer extraction (i.e., producing a valid
permutation), (2) the dense reward, and (3) the sparse reward for Qwen2.5-7B-Instruct-1M.
As shown in Figure~\ref{fig:valid}, all three metrics improve smoothly over training steps, indicating stable optimization without severe oscillation or collapse. 
Notably, the dense reward increases earlier and more steadily
than the sparse reward.
Overall, these validation trends suggest that the proposed reconstruction-based
RLVR framework provides a stable and effective training.

\begin{table*}[ht]
\centering
\small
\setlength{\tabcolsep}{6pt}
\begin{tabular}{c c cc cc cc}
\toprule
\multirow{2}{*}{Model} & \multirow{2}{*}{Task} & \multicolumn{2}{c}{32k} & \multicolumn{2}{c}{64k} & \multicolumn{2}{c}{128k} \\
\cmidrule(lr){3-4}\cmidrule(lr){5-6}\cmidrule(lr){7-8}
 &  & base & ours & base & ours & base & ours \\
\midrule
\multirow{6}{*}{Qwen}
 & vt   & 58.32 & 67.36 & 46.80 & 61.56 & 54.84 & 56.28 \\
 & cwe  & 87.92 & 90.66 & 82.46 & 84.38 & 69.78 & 73.42 \\
 & fwe  & 85.53 & 91.27 & 83.07 & 84.33 & 77.13 & 80.33 \\
 & qa\_1 & 77.40 & 78.40 & 73.00 & 71.20 & 70.60 & 65.00 \\
 & qa\_2 & 60.00 & 64.20 & 56.80 & 61.40 & 49.40 & 59.60 \\
\cmidrule(lr){2-8}
 & avg  & 73.83 & \textbf{78.38} & 68.43 & \textbf{72.57} & 64.35 & \textbf{65.65} \\
\midrule
\multirow{6}{*}{LLaMA}
 & vt   & 78.48 & 83.76 & 54.04 & 77.20 & 24.72 & 49.48 \\
 & cwe  & 86.54 & 92.56 & 60.92 & 80.20 & 10.96 & 28.14 \\
 & fwe  & 81.40 & 87.00 & 73.47 & 73.93 & 61.00 & 67.67 \\
 & qa\_1 & 76.60 & 74.40 & 74.20 & 74.40 & 68.00 & 67.80 \\
 & qa\_2 & 47.80 & 56.20 & 43.20 & 51.40 & 40.00 & 43.00 \\
\cmidrule(lr){2-8}
 & avg  & 74.16 & \textbf{78.78} & 61.17 & \textbf{71.43} & 40.94 & \textbf{51.22} \\
\bottomrule
\end{tabular}
\caption{Performance comparison across sequence lengths.}
\label{tab:qwen_llama_len}
\end{table*}

\subsection{Ruler Score Details}
\label{reuler_app}

As shown in Table~\ref{tab:qwen_llama_len}, we report the evaluation scores for each subtask for RULER.
We can see that our method can surpass base models in almost all cases.

\subsection{Scaling Data to 3W}
\label{scale3w}

We collect 10,000 samples from each domain: books, arXiv, and code.
We end up with 30,000 long document.
The ratio of data with $K = 2,4,6,8$ is $1:1:2:2$.
We train on backbone model Qwen2.5-7B-Instruct-1M.
The performance of our model can surpass baseline by about 10 points.

\end{document}